\documentclass[sn-mathphys]{sn-jnl}

\usepackage{pifont}
\usepackage{multirow}
\usepackage{amssymb}
\newlength\savewidth\newcommand\shline{\noalign{\global\savewidth\arrayrulewidth
\global\arrayrulewidth 1pt}\hline\noalign{\global\arrayrulewidth\savewidth}}

\jyear{2021}%

\theoremstyle{thmstyleone}%
\theoremstyle{thmstyletwo}%

\theoremstyle{thmstylethree}%

\begin{document}

\title[SeaVIS]{SeaVIS: Sound-Enhanced Association for Online Audio-Visual Instance Segmentation}


\author[1,2]{\fnm{Yingjian} \sur{Zhu}}

\author[2]{\fnm{Ying} \sur{Wang}}

\author[1,2]{\fnm{Yuyang} \sur{Hong}}

\author[3]{\fnm{Ruohao} \sur{Guo}}

\author[2]{\fnm{Kun} \sur{Ding}}

\author[5]{\fnm{Xin} \sur{Gu}}

\author[4]{\fnm{Bin} \sur{Fan}}

\author[2,1]{\fnm{Shiming} \sur{Xiang}}

\affil[1]{\orgdiv{School of Artificial Intelligence}, \orgname{University of Chinese Academy of Sciences}, \orgaddress{\city{Beijing} \postcode{101408},  \country{China}}}

\affil[2]{\orgdiv{State Key Laboratory of Multimodal Artificial Intelligence
Systems (MAIS)}, \orgname{Institute of Automation, Chinese Academy of Sciences}, \orgaddress{\city{Beijing} \postcode{100190},  \country{China}}}

\affil[3]{\orgdiv{National Key Laboratory of General Artificial Intelligence, School of Intelligence Science and Technology}, \orgname{Peking University}, \orgaddress{\city{Beijing} \postcode{100871},  \country{China}}}

\affil[4]{\orgdiv{School of Intelligent Science and Technology}, \orgname{University of Science and Technology Beijing}, \orgaddress{\city{Beijing} \postcode{100083},  \country{China}}}

\affil[5]{\orgdiv{Research and Development Department}, \orgname{China Academy of Launch Vehicle Technology}, \orgaddress{\city{Beijing} \postcode{100076},  \country{China}}}


\abstract{Recently, an audio-visual instance segmentation (AVIS) task has been introduced, aiming to identify, segment and track individual sounding instances in videos. However, prevailing methods primarily adopt the offline paradigm, that cannot associate detected instances across consecutive clips, making them unsuitable for real-world scenarios that involve continuous video streams. To address this limitation, we introduce SeaVIS, the first online framework designed for audio-visual instance segmentation. SeaVIS leverages the Causal Cross Attention Fusion (CCAF) module to enable efficient online processing, which integrates visual features from the current frame with the entire audio history under strict causal constraints. A major challenge for conventional VIS methods is that appearance-based instance association fails to distinguish between an object's sounding and silent states, resulting in the incorrect segmentation of silent objects. To tackle this, we employ an Audio-Guided Contrastive Learning (AGCL) strategy to generate instance prototypes that encode not only visual appearance but also sounding activity. In this way, instances preserved during per-frame prediction that do not emit sound can be effectively suppressed during instance association process, thereby significantly enhancing the audio-following capability of SeaVIS. Extensive experiments conducted on the AVISeg dataset demonstrate that SeaVIS surpasses existing state-of-the-art models across multiple evaluation metrics while maintaining a competitive inference speed suitable for real-time processing.}




\maketitle

\section{Introduction}
\label{sec:intro}

\begin{figure*}[htbp]
\centering
\includegraphics[width=\columnwidth]{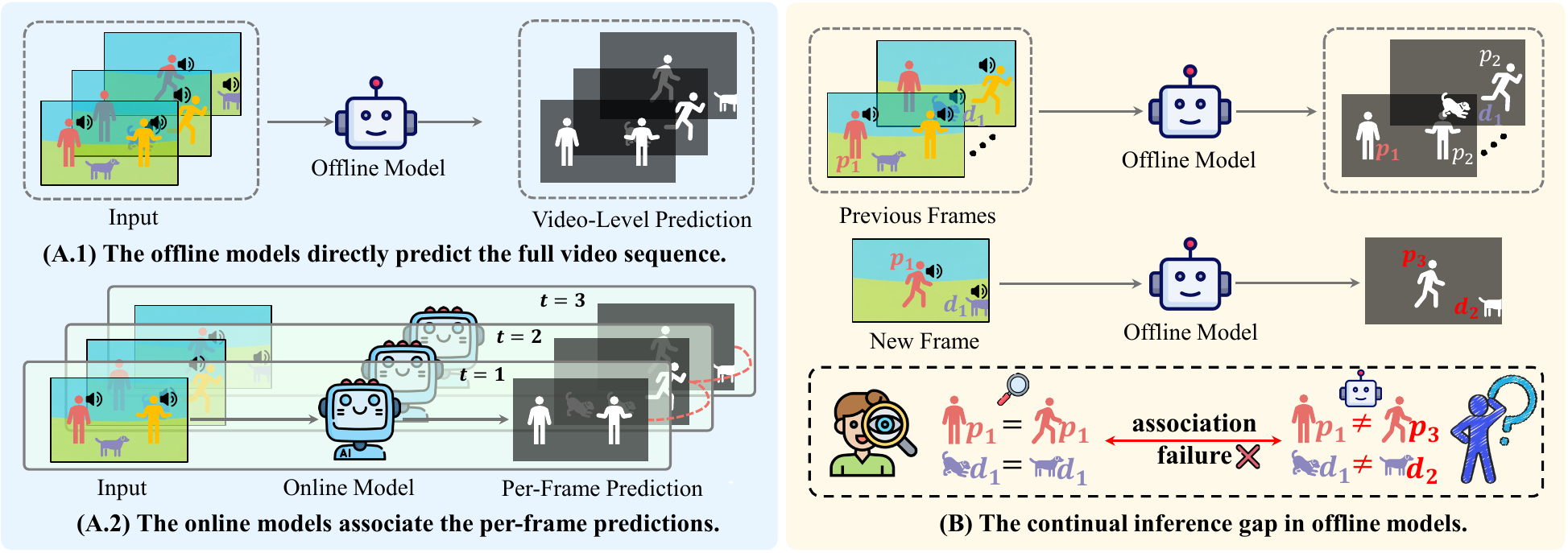}
\caption{Two key limitations of offline models. (A) Different inference paradigms between offline and online models. (A.1) Offline models process the entire sequence of video frames simultaneously and generate only video-level predictions. Consequently, the segmentation of each frame depends heavily on information from future frames. (A.2) In contrast, online models process input frames in a streaming, frame-by-frame manner, followed by post-processing to associate instances across frames. (B)  Offline models face the continual inference gap problem: after completing inference on a fixed-length video segment, the model cannot incrementally process new frames. Consequently, newly predicted segments cannot be associated with previously identified instances (e.g., $p_1$ and $d_1$), leading to association failure such as $p_1 \ne p_3$ and $d_1 \ne d_2$.}
\label{fig:teaser}
\end{figure*}

Audio-Visual Segmentation (AVS)~\cite{avss} has recently emerged as a crucial technology in multi-modal perception, supporting various downstream applications~\cite{mib, xmwang}. However, conventional AVS methods are constrained to the semantic level and cannot distinguish between multiple instances within the same category. In contrast, humans exhibit a remarkable ability to associate distinct sounds with their corresponding visual instances, even within complex and noisy environments. For example, when a group of people is speaking, human can locate the sound sources and estimate the number of individuals who are talking. To address this limitation, AVIS~\cite{avis} introduced the Audio Visual Instance Segmentation task and proposed a baseline model. This model follows the pipeline of VITA~\cite{vita} and employs a window-based attention mechanism to integrate audio-visual temporal information, achieving strong performance on the AVIS test set.

As shown in Fig.~\ref{fig:teaser}, existing methods typically adopt an offline framework which presents two major limitations. First, as illustrated in Fig.~\ref{fig:teaser} (A.1), offline models must process the entire sequence of input frames simultaneously and generates only video-level predictions. As a result, the segmentation output for each frame is highly dependent on information from future frames. Second, offline models exhibit a ``continual inference gap'' (Fig.~\ref{fig:teaser} (B)): once inference is completed on a fixed-length video segment, the model cannot incrementally process newly arriving frames, as the newly predicted segments cannot be associated with previously segmented instances. Moreover, since offline methods can process only video segments of predetermined length, they are ill-suited for continuous streaming visual data encountered in real-world applications.

In contrast, online model processes each frame individually and sequentially, associating instances across frames via post-processing as shown in Fig.~\ref{fig:teaser} (A.2). This online paradigm enables the processing of continuous video streams of arbitrary length, showing a critical capability for applications demanding real-time performance. Therefore, designing an online AVIS framework is essential for its significant practical value in real-world scenarios that require immediate processing and response.

A primary challenge in designing such an online framework for AVIS is to develop an effective fusion method between visual and audio features under strict temporal constraints. Prevailing AVS methods~\cite{combo, avsegformer, hao2024improving} often utilize an ``in-frame fusion" strategy, which overlook the rich temporal information within the audio modality due to the single frame audio-visual aggregation. Although computationally straightforward, this approach inherently limits the model's capacity to capture sequential dependencies over time, thereby reducing its robustness in complex real-world scenarios. In contrast to visual inputs, which typically convey detailed and concrete spatial information in each frame, short audio segments are inherently more ambiguous and susceptible to background noise or overlapping sources. Such ambiguity may introduce misleading cues if treated in isolation, resulting in inaccurate or unstable segmentation. Consequently, a more principled design is required—one that preserves the causal nature of online processing while effectively leveraging the rich temporal information accumulated in the audio stream.

Another challenge for online AVIS is to learn task-specific instance representations. Conventional online video instance segmentation methods~\cite{ctvis, idol} mainly focus on how to learn appearance-based instance embedding so that the identity of the ROI object can be maintained continuously across frames. Therefore, directly applying these methods to AVIS may result in instance prototypes failing to accurately capture the acoustic characteristics of each instance. Consequently, relying solely on visual similarity for instance association introduces significant noise into the final output, often leading to false associations or the inclusion of silent instances in the segmentation results. Addressing this issue requires the design of representations that are not only discriminative in appearance but also sensitive to the dynamic vocalization states of each instance.

To address the challenges mentioned above, we propose SeaVIS (\textbf{S}ound-\textbf{E}nhanced \textbf{A}ssociation for Online Audio-\textbf{V}isual \textbf{I}nstance \textbf{S}egmentation), a novel online framework for audio-visual instance segmentation. To the best of our knowledge, our method is the first to achieve the online audio-visual instance segmentation. To effectively leverage the rich contextual information of audio, we propose the Causal Cross Attention Fusion (CCAF) module, which integrates visual features from the current frame with the entire audio history. Moreover, an Audio-Guided Contrastive Learning (AGCL) strategy is proposed to enhance the association of sounding instances while filtering out the preserved silent instances after per-frame predictions.
The main contributions can be summarized as follows:
\begin{itemize}
    \item A Causal Cross Attention Fusion (CCAF) module is proposed, which effectively integrates historical audio temporal information into multi-scale visual features through cross-attention mechanism.
    \item A novel Audio-Guided Contrastive Learning (AGCL) strategy is introduced at frame and instance levels, driving the model to learn instance embedding that encode both appearance information and vocalization states.
    \item To the best of our knowledge, SeaVIS is the first online framework specifically designed for Audio-Visual Instance Segmentation, which achieves efficient and accurate real-time segmentation, achieving state-of-the-art performance on AVISeg test set.
\end{itemize}

\section{Related Works}
\label{sec:rw}
\subsection{Online Video Instance Segmentation}
Early approaches for online VIS primarily followed a tracking-by-detection paradigm. For instance, MaskTrack R-CNN~\cite{masktrackrcnn} extended Mask R-CNN~\cite{maskrcnn} by adding a separate tracking head to associate instances using handcrafted heuristics. Similarly, CrossVIS~\cite{crossvis} utilized features from past frames to guide the segmentation of the current one. MTNet ~\cite{zhuge2024learning} further emphasizes the integration of motion and temporal cues to enhance the robustness of object localization and tracking. The advent of query-based architectures~\cite{deformabledetr, detr, mask2formerforvis} marked a significant shift, enabling instance tracking by associating learnable query embeddings over time. Within this paradigm, IDOL~\cite{idol} introduced a contrastive learning objective within two adjacent frames to learn more discriminative embeddings for association. However, IDOL exhibits a discrepancy between its training and inference processes. Our method builds upon CTVIS~\cite{ctvis}, a framework developed to address this inconsistency by enhancing the robustness and alignment of the contrastive learning process with the inference pipeline, thereby achieving further substantial performance gains. TCOVIS~\cite{tcovis} enforces long-range consistency through a global instance assignment strategy during training and a spatio-temporal enhancement module. VISAGE~\cite{visage} identifies a specific failure mode in existing query-based methods—an over-reliance on positional information.

\begin{figure*}[t]
\centering
\includegraphics[width=\columnwidth]{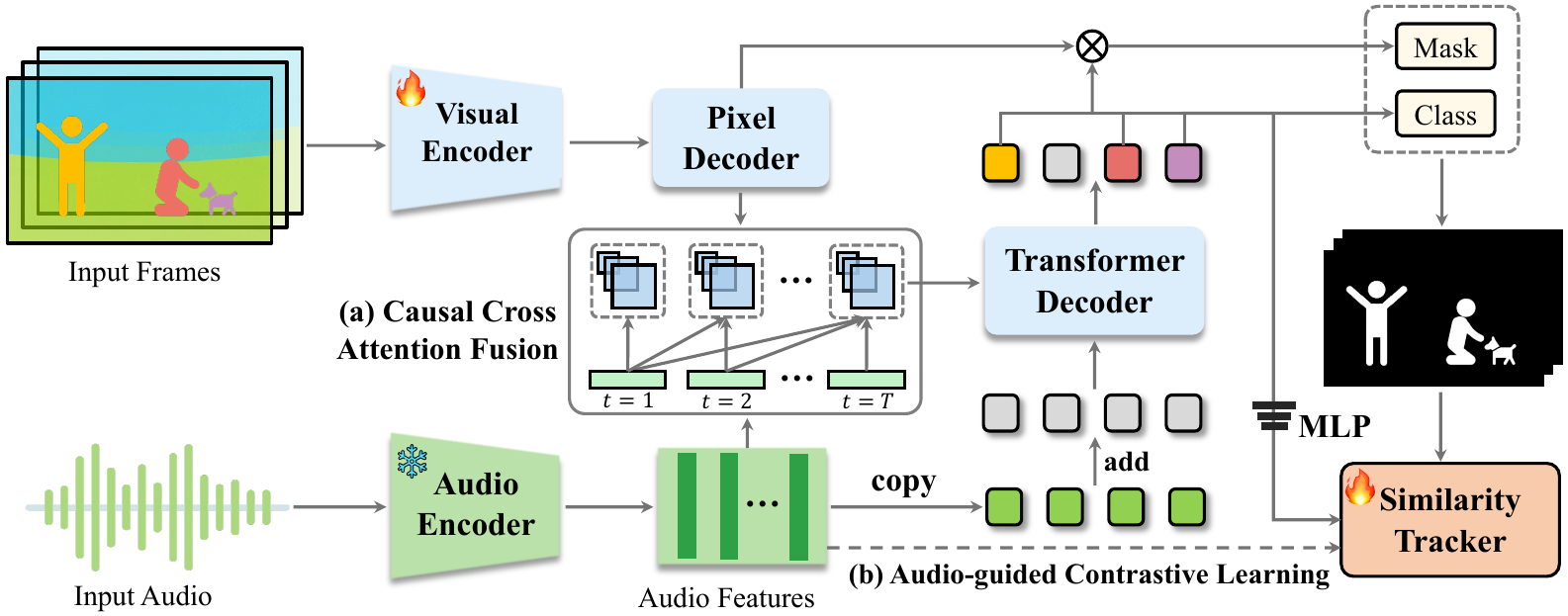}
\caption{Overview of the proposed SeaVIS. SeaVIS operates through two sequential stages: per-frame instance segmentation prediction followed by cross-frame instance association. We propose two key components in our framework: (a) A causal cross attention fusion module enabling cross-frame integration between visual and audio modalities under the temporal constraints of online processing; (b) Dual-level audio-guided contrastive learning at both frame and instance levels to optimize sound-aware instance embedding for audio-visual instance segmentation.}
\label{fig:overview}
\end{figure*}

\subsection{Offline Video Instance Segmentation}
Offline VIS methods have evolved rapidly~\cite{maskprop, vistr, ifc, seqformer, vita}, with each generation introducing new abstractions to better capture spatio-temporal context. MaskProp~\cite{maskprop} extended Mask R-CNN by introducing mask propagation from a reference frame to the entire clip. VisTR~\cite{vistr} pioneered the use of Transformers, reframing VIS as an end-to-end sequence prediction problem. IFC~\cite{ifc} addressed VisTR’s inefficiency by introducing memory tokens for compact inter-frame communication. SeqFormer~\cite{seqformer} proposed video-level instance queries combined with dynamic frame-level box queries for more robust tracking. Vita~\cite{vita} decoupled detection and association via object tokens, enabling scalable and efficient video-level reasoning.

\subsection{Audio-Visual Segmentation}
Sound source localization aims to estimate the position of a sound source within a video sequence and is considered closely related to the task of audio-visual segmentation (AVS)~\cite{ssl1, ssl2}. Recently, proposal-based paradigms~\cite{xuan2022proposal, xuan2024robust} have emerged to address the limitations of coarse map-based approaches, advocating for direct semantic object-level localization through unsupervised spatial constraints. While localization focuses on spatial estimation, AVS extends this goal by predicting pixel-level masks of sounding objects across video frames. To facilitate progress in this direction, AVSBench~\cite{avsbench, avss} introduced a comprehensive benchmark and proposed the temporal pixel-wise audio-visual interaction module (TPAVI), which injects audio semantics as guidance for visual segmentation. This benchmark has since become a cornerstone for evaluating AVS methods.

Following the rapid success of query-based architectures in vision-language tasks, a growing number of AVS approaches have embraced this paradigm~\cite{avsegformer, CATR, hao2024improving, qdformer, stepstones, liu2024annotation}. For example, Avsegformer~\cite{avsegformer} combines dense audio-conditioned visual mixing with sparse query-based decoding, thereby aligning audio and visual cues more effectively. Similarly, CATR~\cite{CATR} employs memory-efficient decoupled encoding and audio-constrained queries to ensure that the segmentation process adheres strictly to audio signals, even when visual information is ambiguous. Beyond Transformer architectures, AVS-Mamba~\cite{mamba} introduces a selective state space model to capture long-range dependencies with linear computational complexity. In addition, QDFormer~\cite{qdformer} applies quantization-based decomposition with a shared codebook to disentangle noisy, multi-source audio, enabling more reliable cross-modal alignment. Several studies investigate to incorporate prior knowledge, such as optical flow and SAM~\cite{sam}, to improve performance~\cite{liu2024audio, combo}. For instance, BAVS~\cite{bavs} leverages multi-modal foundation knowledge to filter off-screen noise and establish explicit audio-visual correspondences. CAVP~\cite{cavp} proposes a novel informative sample mining method for contrastive learning to enhance cross-modal understanding. Similarly, CCFormer~\cite{gong2025complementary} proposes a bi-modal contrastive learning framework to promote the alignment of audio-visual features in a unified space. Moreover, information-theoretic approaches based on the Information Bottleneck (IB) principle have been explored to learn minimal sufficient multi-modal representations, effectively suppressing redundancy and enhancing robustness in audio-visual tasks~\cite{li2023audio, yang2024self, mai2022multimodal}. Additionally, some research focuses on refining cross-modal alignment~\cite{liu2025dynamic, liu2025robust}. For instance, RAVS~\cite{liu2025dynamic} improves audio semantic representations to alleviate feature confusion caused by mixed sound sources.

Recently, more challenging related tasks have been proposed to extend the boundaries of AVS. Referring AVS~\cite{refavs} introduces natural language as an additional input, tasking the model with segmenting a specific object based on a descriptive prompt that may contain both audio and visual cues. Another important extension is the Open-Vocabulary AVS, which aims to segment objects from unseen categories during training, thereby testing the generalization capabilities of the model in open-world scenarios~\cite{ovavs}. The focus of this work lies Audio-Visual Instance Segmentation~\cite{avis}, which requires models to not only segment but also identify and track all individual sounding object instances throughout a video.

\subsection{Contrastive Learning}
Contrastive learning has achieved significant progress in recent years~\cite{simclr, moco, clip, dino, wu2022efficient, simcse}. Early frameworks such as MoCo~\cite{moco} and SimCLR~\cite{simclr} demonstrated the effectiveness of contrastive objectives for image-level self-supervised training, enabling the learning of strong feature representations that transfer well to downstream tasks. Building on this foundation, contrastive methods have been successfully extended to a variety of domains, including vision-language pretraining~\cite{clip}, object detection~\cite{dino}, video instance segmentation~\cite{wu2022efficient}, and natural language processing~\cite{simcse}. Specific to the audio-visual domain, robust strategies such as Active Contrastive Set Mining (ACSM)~\cite{xuan2022active} have been introduced to mitigate faulty negatives inherent in random sampling by actively selecting informative samples for discrimination. Inspired by these advancements, we adopt a contrastive learning framework to learn discriminative embeddings between vocalized and silent instances.

\section{Methodology}
The overall framework of our online audio-visual instance segmentation model, termed as SeaVIS, is shown in Fig.~\ref{fig:overview}. For the $t$-th input frame $x_t \in \mathbb{R}^{H\times W \times 3}$, where $H$ and $W$ are the height and width of image, a visual encoder is employed to extract the multi-scale features. The visual features are further processed by a pixel decoder to generate multi-scale per-pixel visual embeddings $P_{i} \in \mathbb{R}^{\frac{H}{2^{i+1}} \times \frac{W}{2^{i+1}} \times C} $, $ i \in \{2, 3, 4\} $, where $C$ is the visual feature dimension. Following the AVSS~\cite{avss}, we adopt VGGish~\cite{vggish} to extract the audio feature $F_a \in \mathbb{R}^{1\times D}$ from the corresponding audio clip, where $D$ is the audio feature dimension. 

The audio feature is then linearly projected to match the visual embedding dimension, resulting in $F_a' \in \mathbb{R}^{1\times C}$. These visual embeddings $P_{i}$ and the projected audio feature $F_a'$ are input to the Causal Cross-Attention Fusion (CCAF) module to produce audio-enhanced visual representations. The transformer decoder subsequently processes these audio-enhanced features along with a set of learnable queries that have been summed with $F_a'$ to generate segmentation results for sounding instances in this frame. Finally, the decoder's output is passed through a Multilayer Perceptron (MLP) to obtain the final instance embeddings, which are used by a similarity-based tracker to associate the current instance with those from previous frames.

\begin{figure}
\centering
\includegraphics[width=0.6\columnwidth]{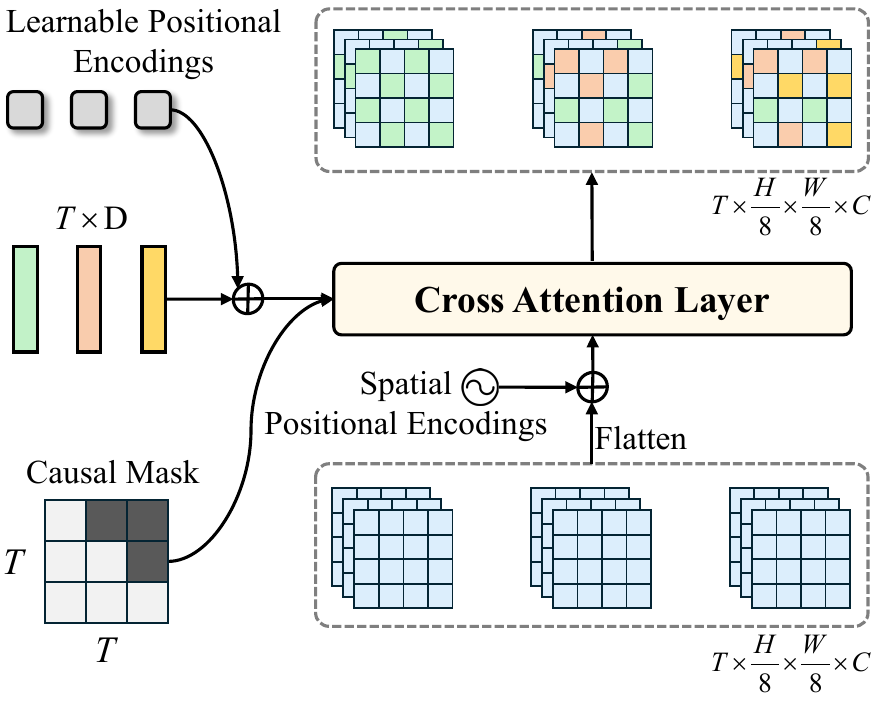}
\caption{Illustration of Causal Cross Attention Fusion(CCAF). This module receives both visual and audio signals, and output the visual feature which is enhanced by audio features. At each frame, the visual features can integrate audio features from all previous and current frames using cross-attention.
}
\label{fig:ccaf}
\end{figure}

\subsection{Causal Cross-Attention Fusion}
Current AVS architectures often rely excessively on visual features while insufficiently utilizing audio signals. For instance, several studies~\cite{avesformer, stepstones} incorporate audio cues merely as weak guidance for query initialization, thereby limiting their contribution to the overall segmentation process. While some Audio-Visual Segmentation (AVS) methods proposed fusion modules, they often adopt the in-frame fusion strategy, neglecting the rich temporal context across consecutive frames within the audio modality. Relying solely on instantaneous audio-visual correspondence is highly restrictive and can easily lead to ambiguous or erroneous interpretations of a scene. To address this limitation, we introduced the Causal Cross Attention Fusion (CCAF) module to facilitate temporal audio-visual interactions. Moreover, to maintain consistency with the principles of an online framework, CCAF employs a causal mask that prevents access to future information, thereby ensuring that the model operates under realistic streaming conditions.

The Causal Cross-Attention Fusion (CCAF) module receives a collection of multi-level, high-resolution per-pixel visual embeddings $P_i \in \mathbb{R}^{T \times \frac{H}{2^{i+1}} \times \frac{W}{2^{i+1}} \times C} $, where $ i \in \{2, 3, 4\} $ alongside the corresponding projected audio feature $F_a'$ as input. Since the audio-visual fusion process is identical across all feature levels, we illustrate the procedure using the highest-resolution pixel-level embedding $P_{2}$, as depicted in Figure~\ref{fig:ccaf}. For visual features, we treat them as a spatiotemporal sequence. Specifically, we flatten the spatial and temporal dimensions and adjust the dimension order to form a query sequence $Q_{in} \in \mathbb{R}^{(T \cdot H' \cdot W') \times C}$. Each element in this sequence represents a pixel-level visual embedding at a specific time and spatial location. 

To enhance the model's ability to recognize sequence order and spatial positions, we introduce positional encodings specific to each modality, following the approach proposed in COMBO~\cite{combo}. Specifically, the visual features incorporate fixed sine spatial positional encodings, whereas the audio features employ learnable positional encodings. Moreover, we utilize a Causal Attention Mask to enforce causality, denoted as $ M \in \mathbb{R}^{(T \cdot H' \cdot W') \times T} $. The mask operates by aligning the flattened visual query sequence with the frame-wise audio key sequence. For any visual embedding at index $i$ in the query sequence, its corresponding timestep is calculated as $t = \lfloor i / (H'W') \rfloor$. The causal mask then ensures that this visual embedding can only attend to audio features from the past and present. The mathematical definition of this mask is:
\begin{equation}
\label{eq:causal_mask}
M_{ij} =
\begin{cases}
0 & \text{if } j \le \lfloor i / (H'W') \rfloor \\
-\infty & \text{otherwise},
\end{cases}
\end{equation}
where $i$ represents the index for each element in the flattened visual query sequence $Q_{in}$ while $j$ is the index for each element in the audio key sequence. This mask is then incorporated into the scaled dot-product attention mechanism before the softmax activation:
\begin{equation}
\label{eq:1}
\text{Attention}(Q, K, V) = \text{softmax}\left(\frac{QK^T}{\sqrt{d_k}} + M\right)V,
\end{equation}
where $d_k$ denotes the dimension of the attention keys. The matrix $Q$ is obtained by linearly projecting the visual features $Q_{\text{in}}$ with the weight matrix $W_Q$, whereas $K$ and $V$ are obtained by projecting the audio features $F_a$ with the weight matrices $W_K$ and $W_V$. The above fusion process is applied independently across all visual feature scales. The CCAF module propagates the rich temporal audio information into the multi-scale visual representations, thereby providing high-quality fused features for the subsequent segmentation.

\subsection{Audio-Guided Contrastive Learning}

\begin{figure*}[t]
\centering
\includegraphics[width=\columnwidth]{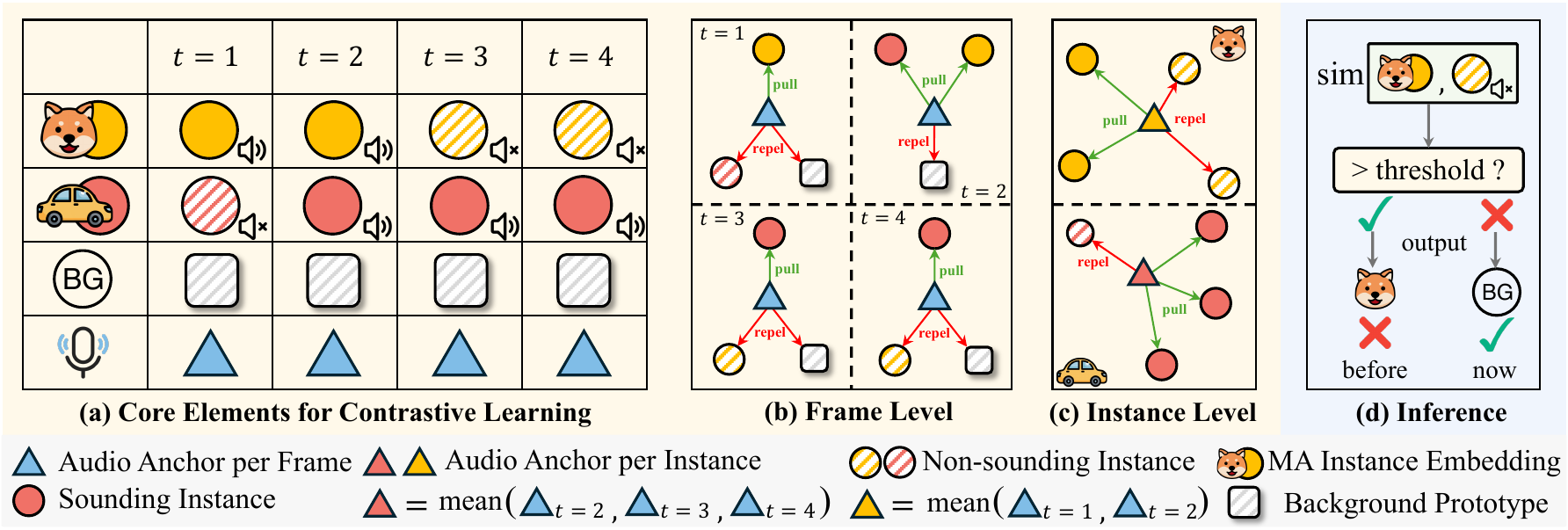}
\caption{Illustration of the Audio-Guided Contrastive Learning (AGCL) strategy. (a) A tracking example contained the essential components of the proposed strategy. At each frame, an anchor is obtained from the corresponding audio clip. The detection results at each time step include sounding instances, non-sounding instances, and background. (b) Frame-level contrastive learning: At each frame, the audio anchor attracts sounding instances while repelling both non-sounding instances and background. (c) Instance-level contrastive learning: For each instance tracked across time, an average audio anchor is computed from frames where the instance is sounding. This anchor pulls the sounding embeddings of that instance while repelling its non-sounding embeddings. (d) The learned sound-aware embeddings assist in filtering out silent instances during the association process.}
\label{fig:agcl}
\end{figure*}

Following per-frame instance segmentation, a tracker module is introduced to associate detection results across frames with instance embeddings derived from the decoder's output using a multilayer perceptron (MLP). Traditional online Video Instance Segmentation (VIS) methods maintain identity consistency across frames by leveraging visual appearance. However, they fail to address a key challenge in Audio-Visual Instance Segmentation (AVIS): per-frame predictions are often coarse and frequently include numerous silent instances. Consequently, post-processing association is required to filter out silent instances and enhance the model's ability to follow audio cues. During inference, online VIS approaches typically adopt a low detection threshold to minimize missed detections and improve mean Average Precision (mAP). This practice, however, allows many silent instances to exceed the threshold and be included in the final output. The underlying cause of this issue lies in the limited discriminative capability of the tracker: different vocal states of an instance cannot be effectively distinguished in the embedding space.

For the reasons mentioned above, we employ contrastive learning to generate instance embeddings that encode not only visual appearance but also an instance's acoustic activity. Specifically, our objectives are twofold: (1) within a single frame, instance embeddings of vocalizing instances should be clearly distinguishable from those of non-vocalizing instances; and (2) for the same instance across different frames, embeddings corresponding to vocalizing and non-vocalizing states should be distinctly separable. These two objectives are designed to be complementary: the first distinguishes between sounding and silent objects within a single frame, while the second separates the sounding and silent states of the same instance across different frames. Therefore, we propose audio-guided contrastive learning strategy at both the frame and instance levels, as illustrated in Fig.~\ref{fig:agcl}.

\noindent \textbf{Frame-level Audio-Visual Contrastive Loss.}
To effectively distinguish sounding instances from silent instances within each frame, we introduce the Frame-level Contrastive Loss($\mathcal{L}_f$). For each frame, an audio anchor $\mathbf{a}_f$ is generated from the corresponding audio clip via a linear layer. The segmentation results in this frame is then partitioned into a positive set, $\mathcal{P}_f$ (all sounding instances) and a negative set, $\mathcal{N}_f$ (all non-sounding instances and background query embeddings). A naive strategy of using anchor $\mathbf{a}_f$ to attract all positive samples in $\mathcal{P}_f$ simultaneously could cause their embeddings to collapse, making them indistinguishable.

To prevent this, we compute the distinct contrastive loss for each sounding instance as illustrated in Fig.~\ref{fig:agcl}(b). Specifically, for any given sounding instance $\mathbf{p}_i \in \mathcal{P}_f$, it serves as the sole positive sample, while all $n_i \in \mathcal{N}_f$ are considered as negative samples during the contrastive loss computation. We use the cosine similarity and introduce a temperature hyperparameter $\tau$ to control the sharpness of the probability distribution. We compute cosine similarity on L2-normalized embeddings. For any vector $\mathbf{x}$, define the normalized vector $\hat{\mathbf{x}} = \mathbf{x} / (\|\mathbf{x}\|_2 + \varepsilon)$, where $\varepsilon = 10^{-12}$ is a small constant added for numerical stability. The loss function follows the InfoNCE-style~\cite{moco} as follows, where $s_{f,i}$ denotes the normalized similarity between all candidate instances in the given frame and the audio anchor:

\begin{equation}
s_{f,i} = \frac{\exp(\hat{\mathbf{a}}_f^\top \hat{\mathbf{p}}_i / \tau)}{\exp(\hat{\mathbf{a}}_f^\top \hat{\mathbf{p}}_i / \tau) + \sum_{\mathbf{n}_j \in \mathcal{N}_f} \exp(\hat{\mathbf{a}}_f^\top \hat{\mathbf{n}}_j / \tau)} ,
\end{equation}

\begin{equation}
\mathcal{L}_{f} = -\frac{1}{\sum_{f} \lvert \mathcal{P}_f \rvert} \sum_{f} \sum_{\mathbf{p}_i \in \mathcal{P}_f} \log(s_{f,i})
\end{equation}

\noindent \textbf{Instance-level Audio-Visual Contrastive Loss.}
This strategy performs contrastive learning on the embeddings of a single tracked instance across all frames within a video clip. The core idea is to create a dedicated audio anchor for each tracked instance $k$, denoted as $\bar{\mathbf{a}}_k$. This anchor represents the instance's unique sound signature and is calculated by averaging the frame-level audio anchors $\mathbf{a_t}$ from a set of frames, $\mathcal{S}_k$, where instance $k$ is sounding. The mean anchor $\bar{\mathbf{a}}_k$ is computed as follows:
\begin{equation}
    \bar{\mathbf{a}}_k = \frac{1}{\lvert \mathcal{S}_k \rvert} \sum_{t \in \mathcal{S}_k} \mathbf{a}_t.
\end{equation}

With the instance-specific anchor, we formulate a multi-positive contrastive learning objective. For each tracked instance $k$, the positive set $\mathcal{P}_k$ comprises its visual embeddings from all sounding frames ($t \in \mathcal{S}_k$), while the negative set $\mathcal{N}_k$ comprises its embeddings from all silent frames ($t \in \mathcal{U}_k$). The final loss is calculated by defining an InfoNCE loss for each eligible instance (those present in both sounding and silent states) and then averaging these losses. The entire process can be captured in the following equation, where $s_{k}$ denotes the normalized similarity between instance $k$ and its corresponding audio anchor, and $N_{inst}$ indicates the number of the eligible instances:

\begin{equation}
s_{k} = \frac{\sum_{\mathbf{p}_i \in \mathcal{P}_k} \exp(\hat{\bar{\mathbf{a}}}_k^\top \hat{\mathbf{p}}_i / \tau)}{\sum_{\mathbf{p}_i \in \mathcal{P}_k} \exp(\hat{\bar{\mathbf{a}}}_k^\top \hat{\mathbf{p}}_i / \tau) + \sum_{\mathbf{n}_j \in \mathcal{N}_k} \exp(\hat{\bar{\mathbf{a}}}_k^\top \hat{\mathbf{n}}_j / \tau)},
\end{equation}
 
\begin{equation} \label{eq:icl_rewrite}
\mathcal{L}_{i} = - \frac{1}{N_{inst}} \sum_{k} \log(s_k).
\end{equation}

\subsection{Training and Inference}
\noindent \textbf{Training}
The overall training loss is computed as follows:
\begin{equation}
\mathcal{L}_{frame} = \lambda_{cls} \mathcal{L}_{cls} + \lambda_{ce} \mathcal{L}_{ce} + \lambda_{dice} \mathcal{L}_{dice},
\end{equation}

\begin{equation}
\mathcal{L} = \mathcal{L}_{frame} + \lambda_{emb} \mathcal{L}_{emb} + \lambda_{f} \mathcal{L}_{f} + \lambda_{i} \mathcal{L}_{i},
\end{equation}
where the per-frame loss $\mathcal{L}_{\text{frame}}$ is responsible for supervising single-frame instance segmentation, following the methodology in Mask2Former~\cite{mask2former}. The embedding loss $\mathcal{L}_{\text{emb}}$, which is identical to that used in CTVIS, serves as a similarity learning objective to facilitate the cross-frame association.

\noindent \textbf{Inference}
The inference pipeline of the proposed SeaVIS is consistent with CTVIS~\cite{ctvis}, which utilizes Mask2Former~\cite{mask2former} to process each frame and incorporates an external memory bank~\cite{idol} to store the states of previously detected instances. These states are composed of classification scores, segmentation masks, and instance embeddings, with the embeddings derived from the decoder's output using a few MLP layers.  Within the memory bank, each tracklet maintains its historical instance embeddings of all previous frames, along with a momentum-averaged instance embedding (referred to as ``MA Instance Embedding" in Fig.~\ref{fig:agcl}).

Subsequently, the embedding of each detected instance is compared with the momentum-averaged embeddings of all tracklets in the memory bank. If the similarity score exceeds the predefined threshold, the new instance is associated with the corresponding existing tracklet. As illustrated in Fig.~\ref{fig:agcl}(d), when a tracked instance becomes silent, the feature separation enforced by AGCL causes its current embedding to exhibit low similarity to the sound-aware MA prototype. Consequently, the similarity score falls below the matching threshold, leading the model to suppress the instance as background rather than erroneously initializing it as a new tracklet. Otherwise, for valid sounding objects, it is initialized as a new tracklet, assigned a unique identity (ID), and added to the memory bank. A key difference between our framework and previous online methods lies in the post-processing stage: beyond instance association, our method is capable of filtering out silent instances. After all video frames have been processed, the model integrates the segmentation masks and classification scores from each tracklet to generate the final video instance segmentation results. We refer the reader to CTVIS~\cite{ctvis} for further implementation details.

\section{Experiments}
\subsection{Dataset and Metrics}
We evaluated SeaVIS on the AVISeg dataset~\cite{avis}, a large-scale benchmark for audio-visual instance segmentation. This dataset contains 926 long-duration videos with an average length of 61.4 seconds, which is considerably longer than previous audio-visual segmentation datasets. These videos are allocated as follows: 616 for training, 105 for validation, and 205 for testing. AVISeg encompasses 26 common sound categories across four dynamic scenarios: "Music," "Speaking," "Machine," and "Animal." It includes frames with silent (6.14\%), single (34.70\%), and multiple simultaneous (59.16\%) sound sources. In total, the dataset comprises 94,074 masks applied to 56,871 annotated frames. 

\begin{table*}[htp]
\caption{Quantitative evaluation of different models from related tasks on the AVISeg test set. The best results are highlighted in bold.}
\label{tab:mr}
\resizebox{\columnwidth}{!}{
\begin{tabular}{c|ccc|ccc|ccccc}
\shline
Task & Model   & Venue  & Audio & FSLA & HOTA & mAP  & FSLAn & FSLAs  & FSLAm &  AssA & DetA \\

\hline
\multirow{5}{*}{VIS} 
& Mask2Former-VIS~\cite{mask2formerforvis}  & CVPR' 22 & \ding{55} & 29.75 & 52.03 & 28.66 & 0.00 & 25.47 & 36.37  & 64.49 & 43.33 \\

& TeViT~\cite{tevit}   & CVPR' 22 & \ding{55} & 32.28 & 53.67 & 31.52 & 0.00 & 28.07 & 39.18  & 65.27 & 45.10 \\

& SeqFormer~\cite{seqformer}   & ECCV' 22 & \ding{55} & 30.32 & 54.32 & 32.79 & 25.03 & 21.76 & 36.46  & 67.25 & 45.23 \\

& VITA~\cite{vita}   & NeurIPS' 22 & \ding{55} & 38.04 & 57.48 & 36.25 & 15.04 & 27.98 & 47.45  & 69.86 & 48.96 \\

& DAVIS~\cite{dvis}   & ICCV' 23 & \ding{55} & 23.99 & 49.12 & 19.83 & 14.61 & 24.83 & 24.69  & 63.51 & 40.11 \\

& LBVQ ~\cite{lbvq}  & TCSVT' 24 & \ding{55} & 34.73 & 56.97 & 36.58 & 27.71 & 29.52 & 38.96  & 68.34 & 48.83 \\

\hline
\multirow{2}{*}{AVSS} 
& AVSegFormer~\cite{avsegformer} & AAAI' 24 & \checkmark & 35.66 & 55.74 & 35.72 & 18.58 & 27.51 & 43.08  & 67.13 & 48.51  \\
& COMBO ~\cite{combo}     & CVPR' 24 & \checkmark & 39.49 & 57.39 & 37.84 & 21.91 & 27.18 & 49.63  & 68.87 & 50.12 \\

\hline
\multirow{2}{*}{AVIS} 
& AVISM~\cite{avis} & CVPR' 25 & \checkmark & 42.78 & 61.73 & 40.57 & 32.22 & 29.83 & 52.40 & 71.15 & 54.97 \\
& \textbf{SeaVIS} & - & \checkmark & \textbf{44.12} & \textbf{63.71} & \textbf{41.23} & \textbf{34.26} & \textbf{31.15} & \textbf{53.66} & \textbf{72.36} & \textbf{57.67} \\
\shline
\end{tabular}}
\vspace{-5pt}
\end{table*}

To comprehensively evaluate performance, we employ three primary metrics: mean Average Precision (mAP)~\cite{map}, Higher-Order Tracking Accuracy (HOTA)~\cite{hota}, and Frame-Level Sound Localization Accuracy (FSLA)~\cite{avis}. Additionally, we utilize five sub-metrics: FSLAn, FSLAs, and FSLAm from FSLA; and AssA and DetA from HOTA.

\subsection{Implementation Details}
Our SeaVIS is trained on 4 NVIDIA A800 GPUs. We employ the Adam optimizer with a learning rate of $1 \times 10^{-4}$ and a weight decay of 0.05. The model is trained for a total of 24,000 iterations with a batch size of 4, with the learning rate reduced at the 6,000 and 20,000 iterations. For each input video, we randomly sample 10 frames from a continuous 20-frame clip. We use clip-level random crop and flip for data augmentation. Following Mask2former~\cite{mask2former}, we utilize the Multi-Scale Deformable Attention Transformer (MSDeformAttn) as our default pixel decoder. Additionally, we adopt the standard transformer decoder architecture with the number of layers $L=3$ (totaling 9 layers) and the number of queries $N_q=100$. The hyperparameters for the loss function are configured as follows: $\lambda_{emb}=2.0$, $\lambda_{cls}=2.0$, $\lambda_{ce}=5.0$, and $\lambda_{dice}=5.0$. Furthermore, the weights for additional loss components are set to $\lambda_{f}=1.0$ and $\lambda_{i}=1.0$, while the temperature  $\tau$ is set to 0.07. We measure FPS on a single NVIDIA A800 GPU with a batch size of 1. Input frames are resized to a shorter side of 360 pixels while maintaining the aspect ratio (up to 1333 pixels), and each video is processed in sequential segments with a fixed temporal window of 5 frames.

\subsection{Main Results}
We conducted our primary experiments on the AVISeg dataset~\cite{avis} to evaluate our method. As AVIS is an emerging task recently~\cite{avis}, we compare SeaVIS against state-of-the-art methods from two highly related fields: Video Instance Segmentation (VIS) and Audio-Visual Semantic Segmentation (AVSS). 

Table~\ref{tab:mr} reports the detailed quantitative results, using three main metrics (FSLA, HOTA, mAP) and five sub-metrics (FSLAn, FSLAs, FSLAm derived from FSLA; AssA, DetA derived from HOTA). For the VIS methods~\cite{mask2formerforvis, tevit, seqformer, vita, dvis, lbvq}, only video frames were used for training, with the audio modality being disregarded. To ensure a fair comparison, all methods utilize a ResNet-50~\cite{resnet} backbone pre-trained on ImageNet~\cite{imagenet}. Notably, our SeaVIS achieves the best performance across all evaluation metrics. Compared to the previous leading model for the AVIS task,  AVISM~\cite{avis}, our results demonstrate improvements, with gains of 1.34 in FSLA, 1.98 in HOTA, and 0.66 in mAP.

\begin{table}[htp]
\begin{center}
\caption{Comprehensive performance comparison between SeaVIS and existing state-of-the-art methods. The best results are highlighted in bold. Inference speed (FPS) is measured on an NVIDIA A800-SXM4-40GB GPU.}
\label{tab:mr2}
\begin{tabular}{cc|ccc|c}
\shline
Method & Backbone  & FSLA & HOTA & mAP & fps \\
\hline
DVIS-online~\cite{dvis}  & Res-50 & 0.03 & 29.97 & 44.29 & 39.57 \\
IDOL~\cite{idol}  & Res-50 & 7.66 & 46.42 & 45.18 & 23.71 \\
GenVIS-online~\cite{genvis}  & Res-50 & 12.31 & 24.97 & 25.16 & 25.6 \\
MinVIS~\cite{minvis}  & Res-50 & 2.45 & 35.13 & 44.66 & \textbf{50.38} \\
\hline
AVISM~\cite{avis}  & Res-50 & 44.42 & 64.52 & 45.04 & 20.46 \\
\textbf{SeaVIS(ours)}  & Res-50 & \textbf{47.09} & \textbf{66.47} & \textbf{46.28} & 34.65 \\
\hline
AVISM~\cite{avis}  & Swin-L & 52.49 & 71.13 & 53.46 & 14.21 \\
\textbf{SeaVIS(ours)}  & Swin-L & \textbf{54.65} & \textbf{73.85} & \textbf{54.29} & \textbf{19.39} \\
\shline
\end{tabular}
\end{center}
\vspace{-5pt}
\end{table}

In addition to segmentation accuracy, we highlight that SeaVIS also delivers highly competitive real-time performance, as detailed in Table~\ref{tab:mr2}. For this analysis, we evaluate using the three main metrics alongside frames-per-second (fps) to measure inference speed, comparing against previous online VIS methods~\cite{dvis, idol, genvis, minvis} and AVISM with COCO~\cite{coco} pre-trained backbones. The table is organized into three groups: the first group presents classic online VIS methods that do not use the audio modality; the second and third groups compare AVIS-specific models using ResNet-50 and Swin-L~\cite{swin} backbones, respectively. When using ResNet-50, it is seen that SeaVIS not only vastly surpasses the online VIS methods in accuracy but also achieves a competitive inference speed of 34.65 fps. Compared to the previous best AVIS model AVISM, our SeaVIS also achieves the highest scores in both accuracy and speed.

\subsection{Ablation Study}
In this section, we conduct ablation studies to verify the effectiveness of each essential design in the proposed SeaVIS. Specifically, ResNet-50 pretrained on COCO is adopted as the backbone to carry out extensive experiments.

\noindent \textbf{Component analysis of SeaVIS.}
To validate the impact of our proposed Causal Cross Attention Fusion Module (CCAF) and Audio-guided Contrastive Learning strategy (AGCL), we first establish a baseline by eliminating both modules. As shown in Table~\ref{tab:ab1}, after introducing the CCAF module to the baseline, all metrics show significant improvement. This is primarily due to the fact that CCAF effectively integrates audio and visual information, providing rich audio temporal context that helps the visual branch better understand the information within each audio clip. Furthermore, upon integrating the AGCL strategy, all metrics are further enhanced. The most substantial gain is seen in the FSLA score, which surges by 3.97 points. This result indicates that our proposed AGCL strategy effectively improves the quality of instance embeddings. During the post-processing stage of instance association, the model leverages these enhanced embeddings, which encode the sounding status of objects, to suppress instances detected in non-sounding frames.
\begin{table}[htp]
\begin{center}
\renewcommand{\arraystretch}{1.1}
\caption{Impact of the Causal Cross Attention Fusion Module (CCAF) and Audio-guided Contrastive Learning strategy (AGCL). Values represent mean $\pm$ standard deviation.}
\label{tab:ab1}
\begin{tabular}{cc|ccc}
\shline
CCAF     & AGCL     & FSLA    & mAP  & HOTA   \\
\hline
            &             & 42.03 $\pm$ 0.14 & 43.93 $\pm$ 0.12 & 63.54 $\pm$ 0.17 \\
\checkmark &             & 43.10 $\pm$ 0.09 & 45.74 $\pm$ 0.16 & 66.03 $\pm$ 0.15 \\
\checkmark & \checkmark & \textbf{47.03} $\pm$ 0.15 & \textbf{46.11} $\pm$ 0.12  & \textbf{66.39} $\pm$ 0.13 \\
\shline
\end{tabular}
\end{center}
\vspace{-2pt}
\end{table}

\noindent \textbf{Impact of Causal Cross Attention Fusion Module.}
As shown in Table~\ref{tab:ab2}, we conduct an ablation study on three different working modes of our CCAF module: audio-to-visual fusion (inject audio feature into visual), visual-to-audio fusion (inject visual feature into audio), and bilateral fusion (audio-to-visual fusion followed by visual-to-audio fusion). The results indicate that the audio-to-visual fusion mode achieves the best overall performance. We attribute the superior performance of the audio-to-visual fusion to the issue of modality imbalance. In visual-to-audio or bilateral fusion, audio information is often overwhelmed by dense visual features during cross-attention. In contrast, audio-to-visual fusion better preserves the audio modality, thereby enhancing the model’s overall detection performance.

\begin{table}[htp]
\begin{center}
\renewcommand{\arraystretch}{1.1}
\caption{Comparisons between different fusion modes.}
\label{tab:ab2}
\begin{tabular}{c|ccc}
\shline
Fusion Mode     & FSLA    & mAP  & HOTA   \\
\hline
audio $\rightarrow$ visual & \textbf{47.09} & \textbf{46.28} & 66.47 \\
visual $\rightarrow$ audio & 45.91 & 45.06 & 64.28 \\
bilateral fusion & 46.17 & 44.89  & \textbf{66.91} \\
\shline
\end{tabular}
\end{center}
\vspace{-2pt}
\end{table}

\begin{figure}[htbp]
\centering
\includegraphics[width=0.6\columnwidth]{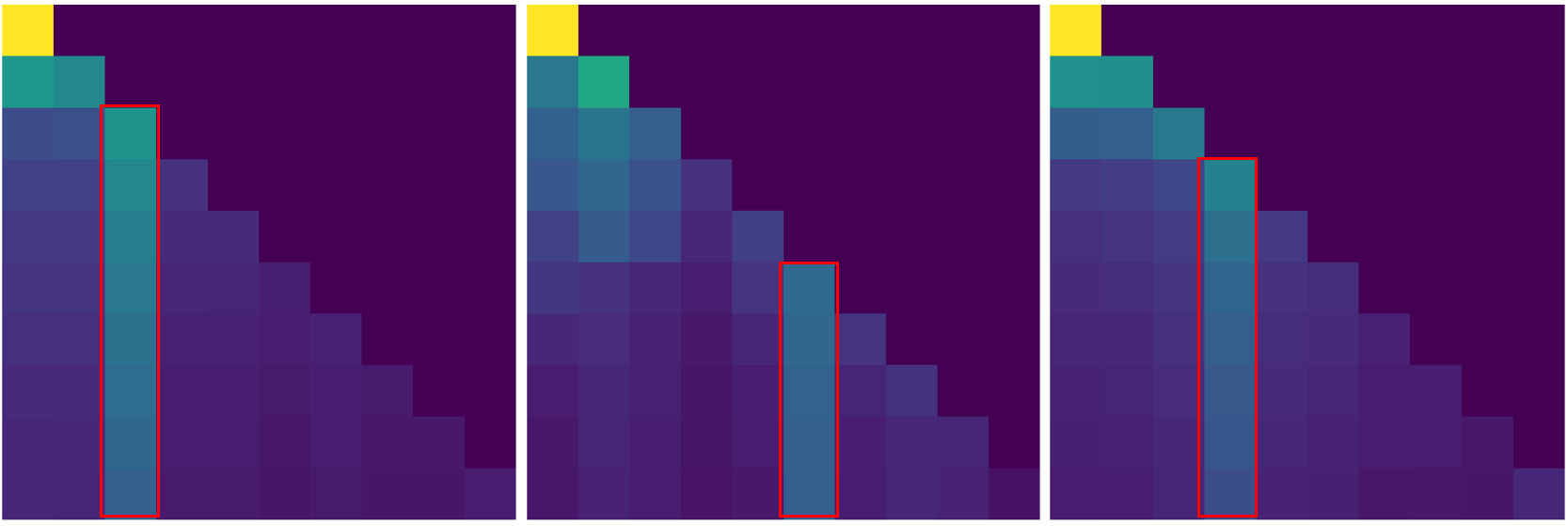}
\caption{Visualization of attention scores from CCAF module during inference. The red boxes highlight that for a given frame, the module assigns high attention scores to critical preceding audio time steps.}
\label{fig:attn}
\end{figure}

As illustrated in Fig.~\ref{fig:attn}, we visualize the attention scores of CCAF during inference, which contains three distinct examples of attention heatmaps. In each map, the score at coordinate $(x,y)$ represents the attention that the visual feature at frame $y$ pays to the audio feature at frame $x$. These heatmaps reveal a consistent pattern: while processing a given frame, the module allocates significantly high attention scores to specific, critical preceding audio frames. These key frames, highlighted with red boxes, often correspond to moments of high informational value, such as the ignition of a lawnmower or the onset of a speech. In parallel, these key historical audio moment can help the current frame better understand the audio clip of the current frame, further demonstrating the effectiveness of our proposed CCAF module.

\noindent \textbf{Impact of Audio-guided Contrastive Learning Strategy.}
As illustrated in Table~\ref{tab:ab3}, we investigate the respective contributions of the two components in our proposed AGCL strategy: the frame-level and instance-level audio-visual contrastive loss. Applying only the frame-level loss substantially improves the FSLA score, yet it degrades the mAP and HOTA metrics. This is because the frame-level loss helps distinguish sounding from silent objects within the same frame. However, it also causes excessive aggregation among different sounding instances, which degrades segmentation and tracking performance. In contrast, although the instance-level loss contributes less to FSLA improvement, it performs well across the other evaluation metrics. Most importantly, when both losses are employed concurrently, the model achieves the best performance on both FSLA and mAP, while maintaining a strong HOTA score. These findings validate the effectiveness of our proposed AGCL strategy, confirming that the frame-level and instance-level work in a complementary way to enhance the model's overall performance.

\begin{table}[htp]
\begin{center}
\caption{Ablation study on Audio-guided Contrastive Learning (AGCL) Strategy.}
\label{tab:ab3}
\renewcommand{\arraystretch}{1.1}
\begin{tabular}{cc|ccc}
\shline
$\lambda_{f}$     & $ \lambda_{i}$      & FSLA    & mAP  & HOTA   \\
\hline
0 &   0   & 43.12 & 45.92 & 66.03 \\
1 &   0   & 45.96 & 45.06 & 65.64 \\
0 &  1    & 44.58 & 46.02 & \textbf{66.97} \\
1 & 1 & \textbf{47.09} & \textbf{46.28}  & 66.47 \\
\shline
\end{tabular}
\end{center}
\end{table}

\noindent \textbf{Effect of Audio Queries.}
As shown in Table~\ref{tab:ab4}, we investigate the effects of different initialization strategies for the learnable queries. Specifically, we compare two approaches: using only audio queries as the initial queries (denoted as "all") and combining audio queries with learnable queries through addition (denoted as "add"). For the "add" strategy, we first expand the audio features to match the dimensions of the learnable queries using  a linear layer. The results clearly indicate that the "add" strategy consistently outperforms the "all" strategy across all evaluation metrics.

\begin{table}[htp]
\begin{center}
\renewcommand{\arraystretch}{1.2}
\caption{Ablation study on query initialization.}
\label{tab:ab4}
\begin{tabular}{c|ccc}
\shline
Queries     & FSLA    & mAP  & HOTA   \\
\hline
\textbf{add} & \textbf{47.09} & \textbf{46.28} & \textbf{66.47} \\
all & 45.91 & 45.06 & 64.28 \\
\shline
\end{tabular}
\end{center}
\vspace{-1pt}
\end{table}

\begin{figure*}[t]
\centering
\includegraphics[width=\columnwidth]{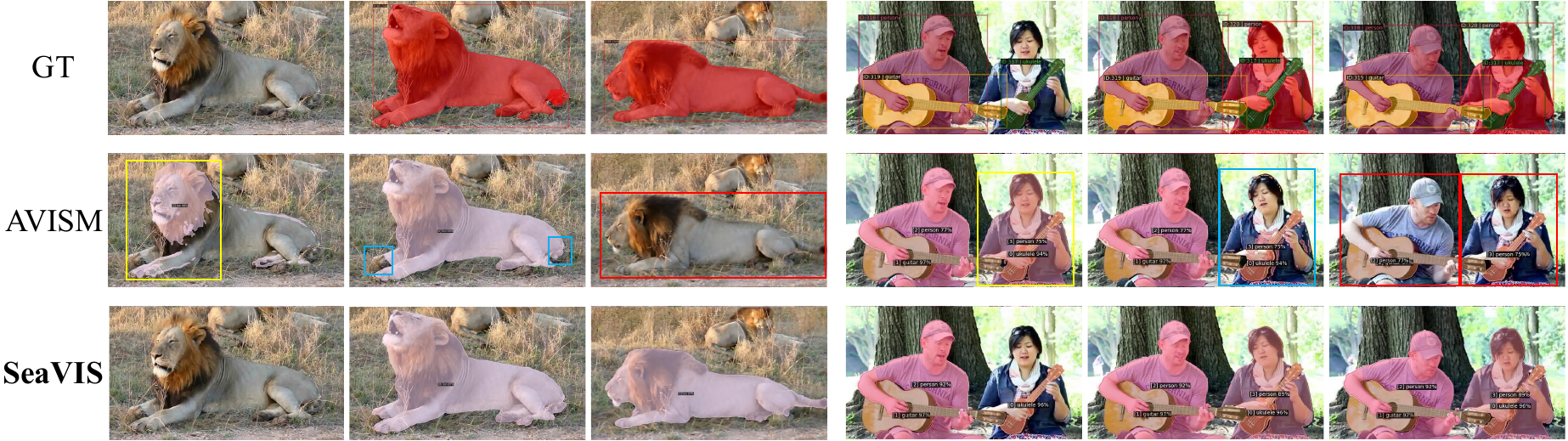}
\caption{Qualitative results of AVISM and SeaVIS on AVISeg test set. These examples show performance under no-sound scene (the left), single sound source (the left) and multi sound sources (the right). The red boxes highlight false negatives (FNs), the yellow boxes highlight false positives (FPs) and blue ones mark inferior segmentations.}
\label{fig:vis}
\end{figure*}

\noindent \textbf{Ablation on Number of Training Frames.}
To verify the effectiveness of long-video training, we ablate the number of sampling frames of each video used for training, denoted as $N_f$. As demonstrated in Table~\ref{tab:ab5}, we experiment with using 2, 5, 10, and 15 frames. The results show a consistent improvement in performance as the number of frames increases from 2 to 10. The model achieves optimal performance with 10 frames. However, when the number of frames is further increased to 15, a slight degradation in performance is observed across all metrics.

\begin{table}[htp]
\begin{center}
\caption{Ablation study on number of training frames.}
\label{tab:ab5}
\renewcommand{\arraystretch}{1.1}
\begin{tabular}{c|ccc}
\shline
$N_f$     & FSLA    & mAP  & HOTA   \\
\hline
2 & 43.85 & 42.69 & 62.91 \\
5 & 45.26 & 44.96 & 64.87 \\
\textbf{10} & \textbf{47.09} & \textbf{46.28} & \textbf{66.47} \\
15 & 46.64 & 45.76 & 66.12 \\
\shline
\end{tabular}
\end{center}
\vspace{-1pt}
\end{table}

\noindent \textbf{Impact of CCAF on Long-video Efficiency.} To explicitly assess the computational overhead introduced by our proposed CCAF module, we conduct a comprehensive analysis of peak GPU memory usage and inference latency on an NVIDIA GeForce RTX 3090. As summarized in Table~\ref{tab:efficiency_tradeoff}, integrating the CCAF module results in negligible overhead across various video durations. Specifically, for short 1-minute videos, the memory footprint increases by approximately 100 MB with a minimal FPS reduction of only 0.11. Even for extended 30-minute streams, memory consumption remains marginal while maintaining real-time performance. This efficiency primarily stems from our segment-based inference strategy, wherein the CCAF module operates within a fixed temporal window rather than computing global attention across the entire video sequence. Theoretically, we posit that overly long temporal contexts are often redundant for resolving local audio-visual ambiguities. These findings demonstrate that SeaVIS achieves substantial accuracy improvements without incurring significant memory or latency costs.

\begin{table}[htp]
    \centering
    \caption{Ablation analysis of the CCAF module’s impact on computational efficiency and inference performance over extended video durations.}
    \label{tab:efficiency_tradeoff}
    {
    \begin{tabular}{lcccccc}
        \toprule
        Duration & CCAF & GPU Mem (MB) & Power (W) & FPS & mAP \\
        \midrule
        \multirow{2}{*}{1 min} & \ding{55} & 5,950 & 224 & 24.23 & 44.19 \\
         & \ding{51} & 6,054 & 228 & 24.12 & 46.28 \\
        \midrule
        \multirow{2}{*}{30 min} & \ding{55} & 20,516 & 228 & 26.58 & 52.18 \\
         & \ding{51} & 21,228 & 225 & 25.49 & 52.96 \\
        \bottomrule
    \end{tabular}
    }
\end{table}

\noindent \textbf{Impact of Audio Backbone and AGCL Stability.} To investigate the performance limits of SeaVIS and verify the robustness of our AGCL strategy across different audio backbones, we evaluate our framework using more advanced audio representations. Specifically, we replace the default VGGish encoder with AudioMAE~\cite{audiomae} and report the results in Table~\ref{tab:audio_backbone}. As observed, adopting a stronger audio backbone yields substantial gains in audio-sensitive metrics, notably boosting FSLA from 47.09 to 49.24 and increasing HOTA from 66.47 to 67.19. Furthermore, we examine whether AGCL remains essential when paired with superior audio embeddings. A comparison between the second and third rows of Table~\ref{tab:audio_backbone} reveals that even with powerful features from AudioMAE, the absence of AGCL leads to a sharp decline in performance, with FSLA dropping significantly from 49.24 to 46.02. These results confirm that the AGCL strategy provides a fundamental and generalizable performance advantage, independent of the capacity of the audio encoder.
\begin{table}[htbp]
    \centering
    \caption{Impact of advanced audio backbones and the consistent effectiveness of the AGCL strategy.}
    \label{tab:audio_backbone}
    {
    \begin{tabular}{lcccc}
        \toprule
        Audio Encoder & AGCL & FSLA & HOTA & mAP \\
        \midrule
        VGGish & \ding{51} & 47.09 & 66.47 & \textbf{46.28} \\
        AudioMAE & \ding{51} & \textbf{49.24} & \textbf{67.19} & 46.19 \\
        AudioMAE & \ding{55} & 46.02 & 66.95 & 45.86 \\
        \bottomrule
    \end{tabular}
    }
\end{table}

{\subsection{Analysis of Robustness and Generalization}

\noindent \textbf{Robustness to Noisy Audio.} To evaluate robustness against audio corruption, we introduced Gaussian white noise ($\mu=1, \sigma^2=0.1$) to the audio stream with a probability of 0.1, as shown in Table~\ref{tab:robustness}. SeaVIS exhibits a more pronounced decline in FSLA and HOTA compared to AVISM. We interpret this sensitivity as a validation of our method’s efficacy: whereas the baseline’s performance invariance suggests a failure to effectively exploit audio dynamics, the observed drop in SeaVIS confirms that our proposed modules actively leverage audio cues for precise localization and association, thereby demonstrating a stronger audio-following capability.

\begin{table}[htp]
    \centering
    \caption{Robustness analysis under noisy audio conditions.}
    \label{tab:robustness}
    {
    \begin{tabular}{llccc}
        \toprule
        Method & Audio & FSLA & HOTA & mAP \\
        \midrule
        AVISM & Clean & 44.42 & 64.52 & 45.04 \\
        AVISM & Noisy & 43.19 \small{($\downarrow$1.23)} & 63.19 \small{($\downarrow$1.33)} & 44.89 \small{($\downarrow$0.15)} \\
        \midrule
        SeaVIS (Ours) & Clean & \textbf{47.09} & \textbf{66.47} & \textbf{46.28} \\
        SeaVIS (Ours) & Noisy & 44.17 \small{($\downarrow$2.92)} & 64.26 \small{($\downarrow$2.21)} & 46.09 \small{($\downarrow$0.19)} \\
        \bottomrule
    \end{tabular}
    }
\end{table}

\noindent \textbf{Tracking Stability under Overlapping Speech.} To rigorously assess the model's tracking stability in complex acoustic environments, we evaluate the Identity Switches (IDSW) metric across varying levels of audio overlap. The AVISeg test set is divided into two subsets based on acoustic complexity: a Single-Source subset and a Multi-Source subset, the latter characterized by overlapping speech from multiple sources. As shown in Table~\ref{tab:idsw_analysis}, SeaVIS consistently achieves lower IDSW scores than AVISM in both scenarios. Notably, in the more challenging Multi-Source setting, SeaVIS reduces the number of identity switches from 71 to 63. This reduction empirically demonstrates SeaVIS's superior association robustness, effectively resolving identity ambiguities even when sound sources overlap spatially or temporally.

\begin{table}[htbp]
    \centering
    \caption{Quantitative analysis of tracking stability (IDSW) under different acoustic complexities.}
    \label{tab:idsw_analysis}
    {
    \begin{tabular}{lcc}
        \toprule
        \multirow{2}{*}{Method} & \multicolumn{2}{c}{IDSW ($\downarrow$)} \\
        \cmidrule(lr){2-3}
         & Single-Source & Multi-Source \\
        \midrule
        AVISM & 17 & 71 \\
        \textbf{SeaVIS (Ours)} & \textbf{10} & \textbf{63} \\
        \bottomrule
    \end{tabular}
    }
\end{table}

\subsection{Qualitative Analysis}
\noindent \textbf{Comparison with AVISM.} We present a qualitative comparison between AVISM~\cite{avis} and our proposed method, SeaVIS, on the AVISeg test set. As depicted in Fig.~\ref{fig:vis}, SeaVIS demonstrates superior spatial localization and audio-following segmentation capabilities, leading to more robust segmentation performance. For example, in the single-source scene (left), SeaVIS accurately distinguishes between the lion's sounding and silent frames. In contrast, AVISM fails to make this distinction, resulting in a false positive (FP) error in the silent frame. Moreover, for the sounding lion, SeaVIS provides a  more precise segmentation mask. We also provide more visual results of our SeaVIS from different scenarios as illustrated in Fig.~\ref{fig:morevis}.

\noindent \textbf{Failure Case Analysis.} We present three representative failure cases in Figure~\ref{fig:failure-case} to analyze the limitations of the system. First, in multi-person scenes involving frequent and rapid turn-taking, the model occasionally struggles to promptly switch the active speaker identity. Second, when vocals and instrumental sounds occur simultaneously, the model may fail to accurately distinguish the specific sound source. Finally, in outdoor environments with low audio capture quality or dominant background noise, the segmentation performance deteriorates. The first two cases stem from limitations in the current audio representation, which requires further enhancement to better disentangle and reference multiple overlapping sound sources. The third case indicates the need for integration of specialized audio pre-processing or denoising techniques.

\begin{figure*}[htbp]
\centering
\includegraphics[width=\columnwidth]{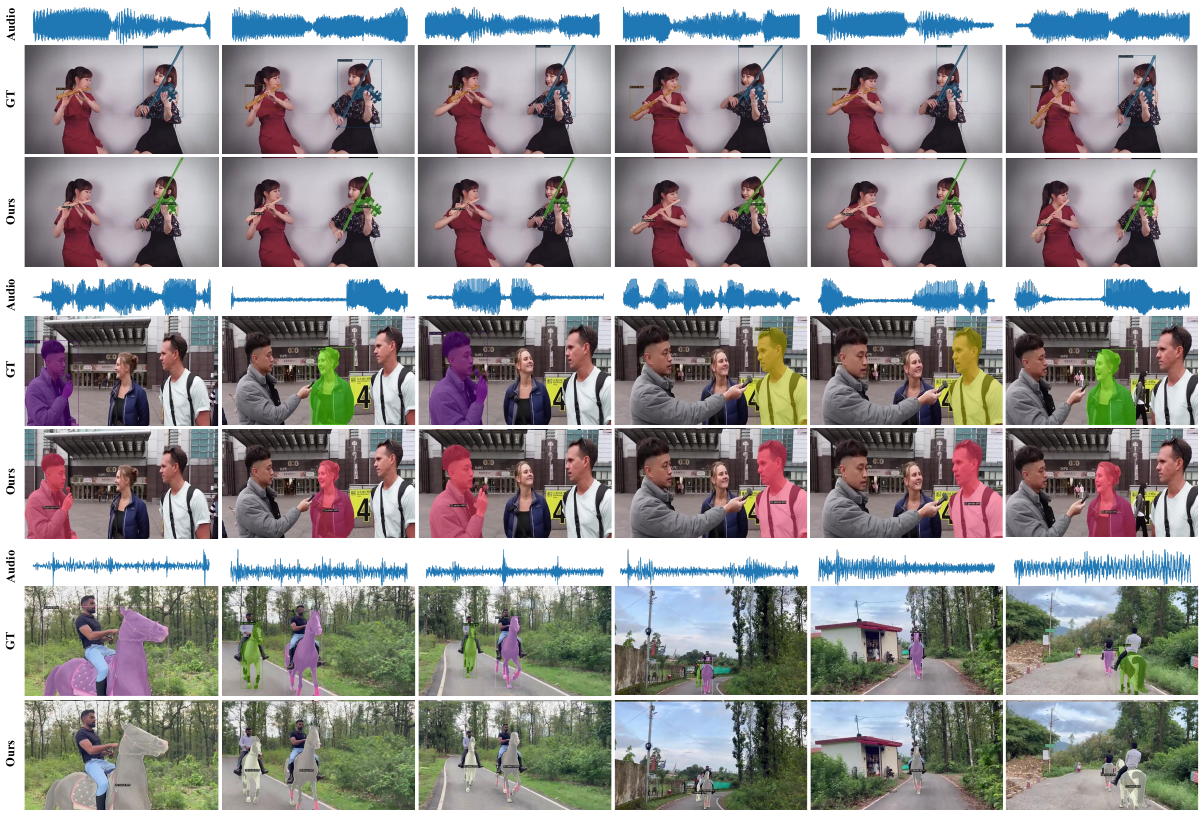}
\caption{More visual results of our SeaVIS on AVISeg dataset from different scenarios. Each row have six sampled frames from a video sequence. Zoom in to see details.}
\label{fig:morevis}
\end{figure*}

\begin{figure}[htbp]
    \centering
    \includegraphics[width=\columnwidth]{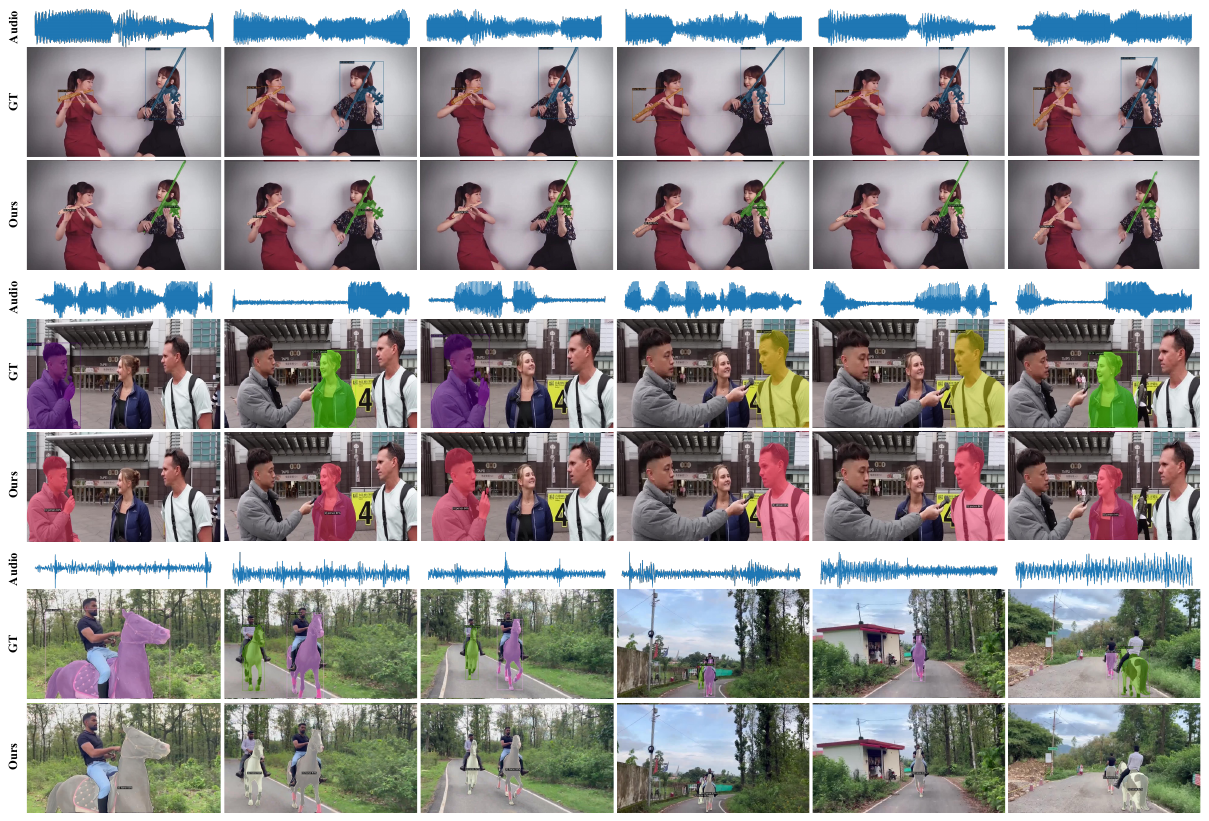}
    \caption{Failure cases of our SeaVIS on AVISeg dataset from different scenarios. Each row have six sampled frames from a video sequence. Zoom in to see details.}
    \label{fig:failure-case}
\end{figure}

\section{Conclusion}
We propose the first online audio-visual instance segmentation framework, named SeaVIS, which achieves state-of-the-art performance on the AVISeg test set. The core of SeaVIS consists of two novel components: the Causal Cross-Attention Fusion (CCAF) module, which effectively integrates historical audio temporal information into multi-scale visual features through a cross-attention mechanism; and the Audio-Guided Contrastive Learning (AGCL) strategy, which encourages the model to learn instance embeddings that encode both visual appearance and vocalization states. Extensive experiments demonstrate that our method not only outperforms existing state-of-the-art models across multiple evaluation metrics but also exhibits efficient real-time processing capabilities. Moreover, our approach is well-suited for streaming video input in real-world scenarios, thereby enabling the practical deployment of audio-visual instance segmentation (AVIS) techniches. This advancement supports many applications in complex environments, such as autonomous driving, interactive robotics, and human-computer interaction.
\newpage


\newpage

\begin{figure}[h]%
\centering
\includegraphics[width=0.3\textwidth]{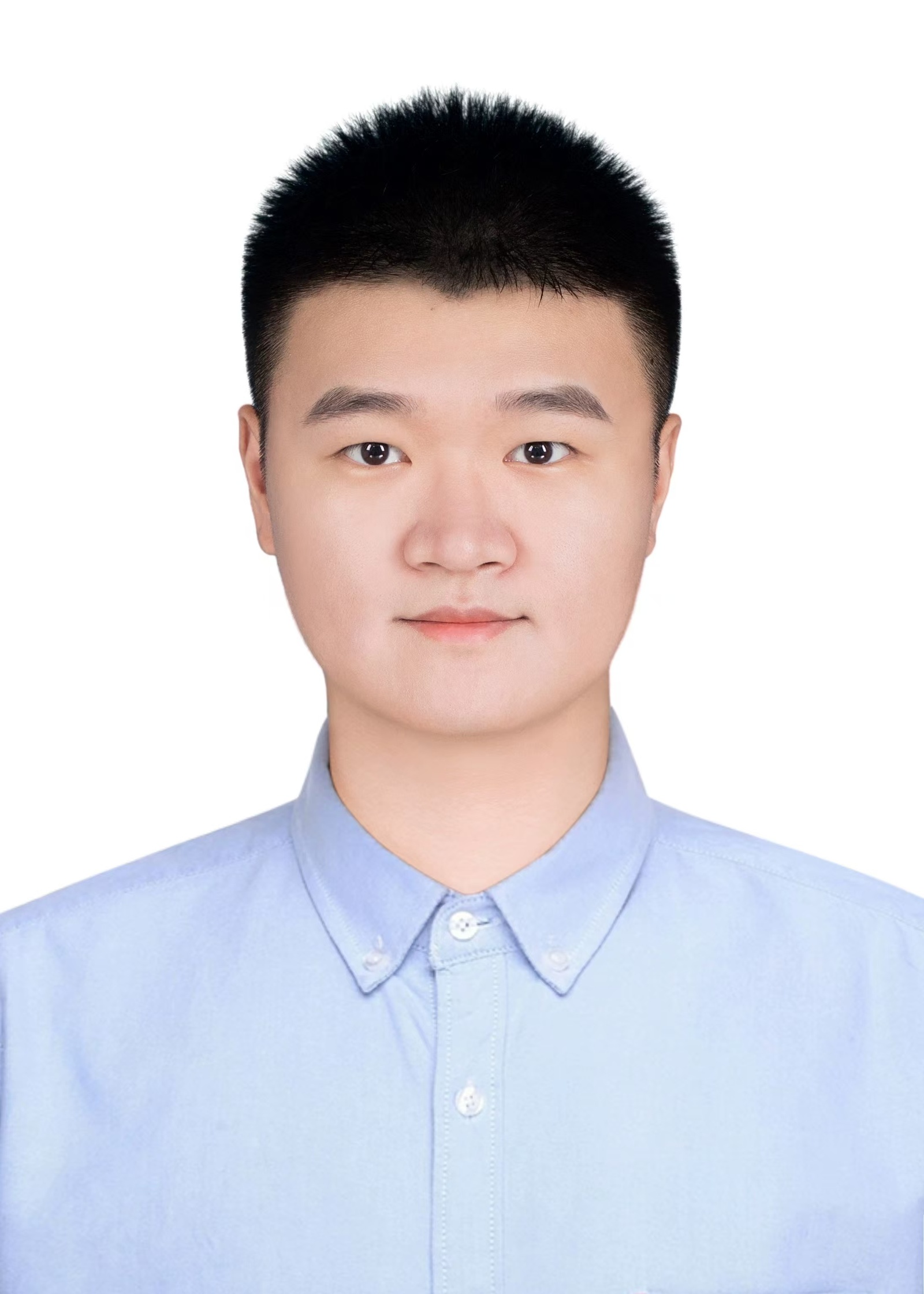}
\vspace{-10pt}
\end{figure}

\noindent{\bf Yingjian Zhu} received his B.S. degree from the Harbin Institute of Technology, Weihai, China, in 2024. He is currently pursuing the Ph.D. in the State Key Laboratory of Multimodal Artificial Intelligence Systems at the Institute of Automation, Chinese Academy of Sciences, and the School of Artificial Intelligence, University of Chinese Academy of Sciences, Beijing, China. 

His current research interests include multi-modal learning, information retrieval and computer vision.

E-mail: zhuyingjian2024@ia.ac.cn

ORCID iD: 0009-0003-5538-0645

\begin{figure}[h]%
\centering
\includegraphics[width=0.3\textwidth]{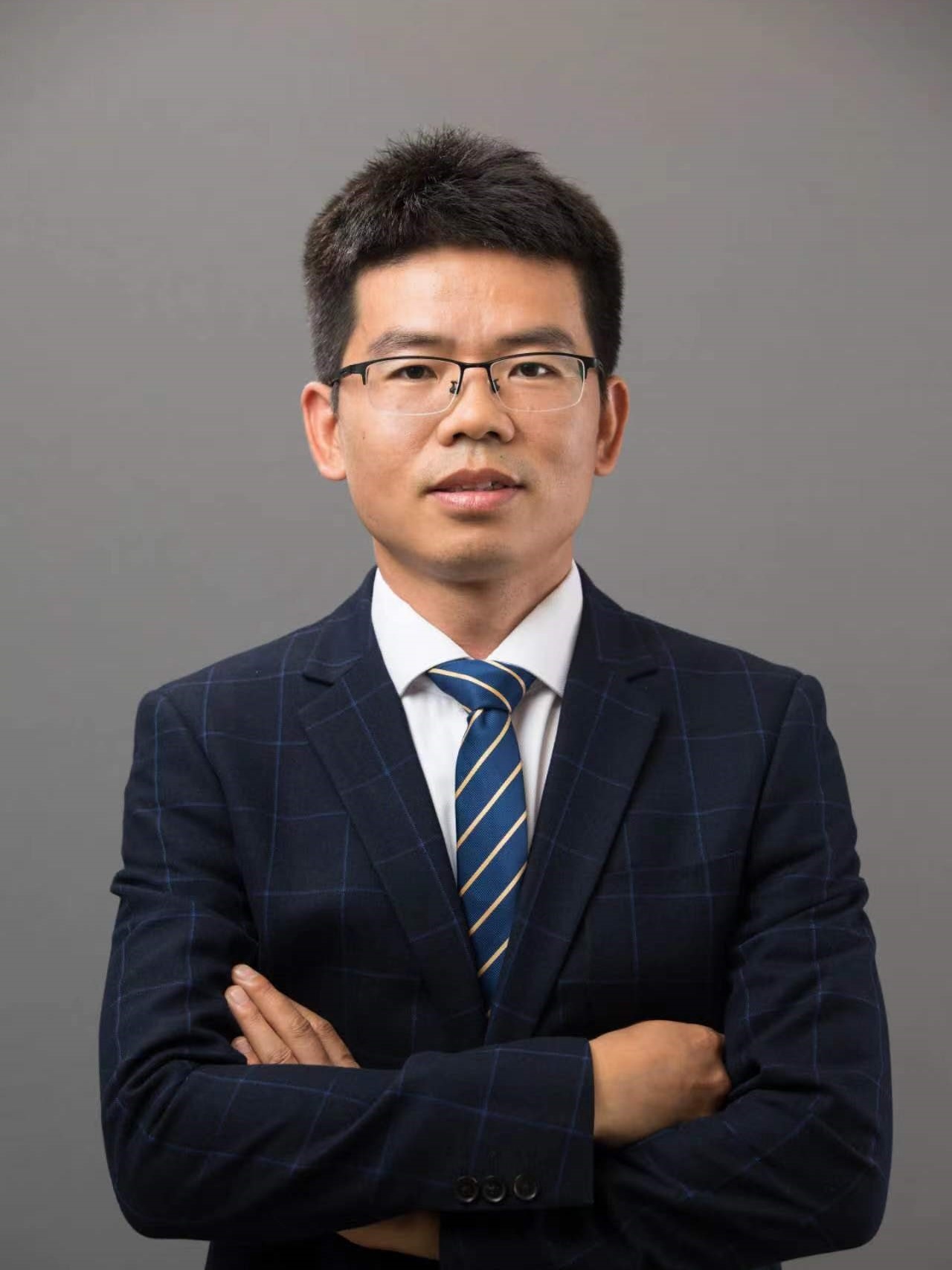}
\vspace{-10pt}
\end{figure}

\noindent{\bf Ying Wang} received the B.S. degree from the Nanjing University of Information Science and Technology, China in 2005, the M.S. degree from the Nanjing University of Aeronautics and Astronautics, China, in 2008, and the Ph.D. degree from the Institute of Automation, Chinese Academy of Sciences, Beijing, China, in 2012. He is currently an Associate Professor with the State Key Laboratory of Multimodal Artificial Intelligence Systems, Institute of Automation, Chinese Academy of Sciences. 

His research interests include computer vision, pattern recognition and remote sensing image processing.

E-mail: ywang@nlpr.ia.ac.cn

\begin{figure}[h]%
\centering
\includegraphics[width=0.3\textwidth]{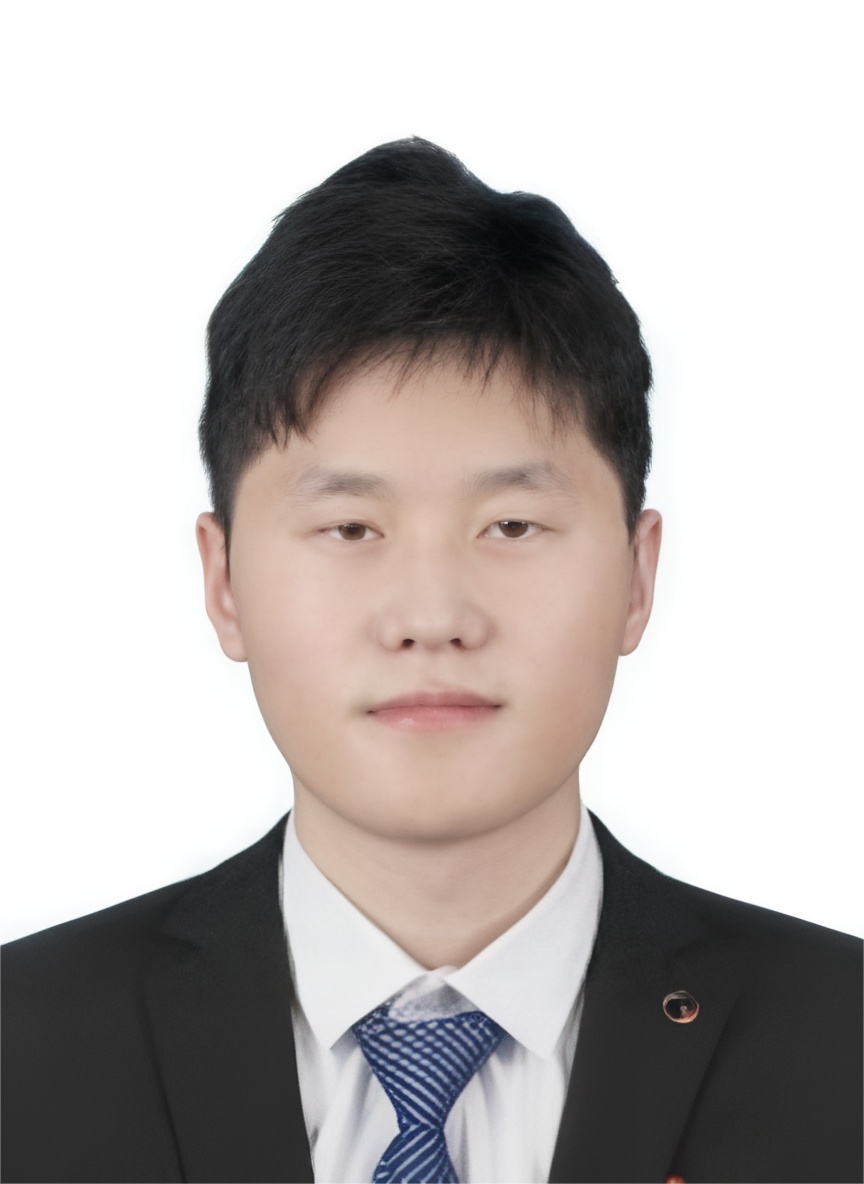}
\vspace{-10pt}
\end{figure}

\noindent{\bf Yuyang Hong} recevied his B.S. degree from the Chongqing University, Chongqing, China, in 2023. He is currently pursuing the Ph.D. in the State Key Laboratory of Multimodal Artificial Intelligence Systems at the Institue of Automation, Chinese Academy of Sciences, and the School of Artificial Intelligence, University of Chinese Academy of Sciences, Beijing, China. 

His current research interests include Multi-modal Learning, Continual Learning and Retrieval-Augmented Generation.

E-mail: hongyuyang2023@ia.ac.cn

ORCID iD: 0009-0000-7281-8272

\begin{figure}[h]%
\centering
\includegraphics[width=0.3\textwidth]{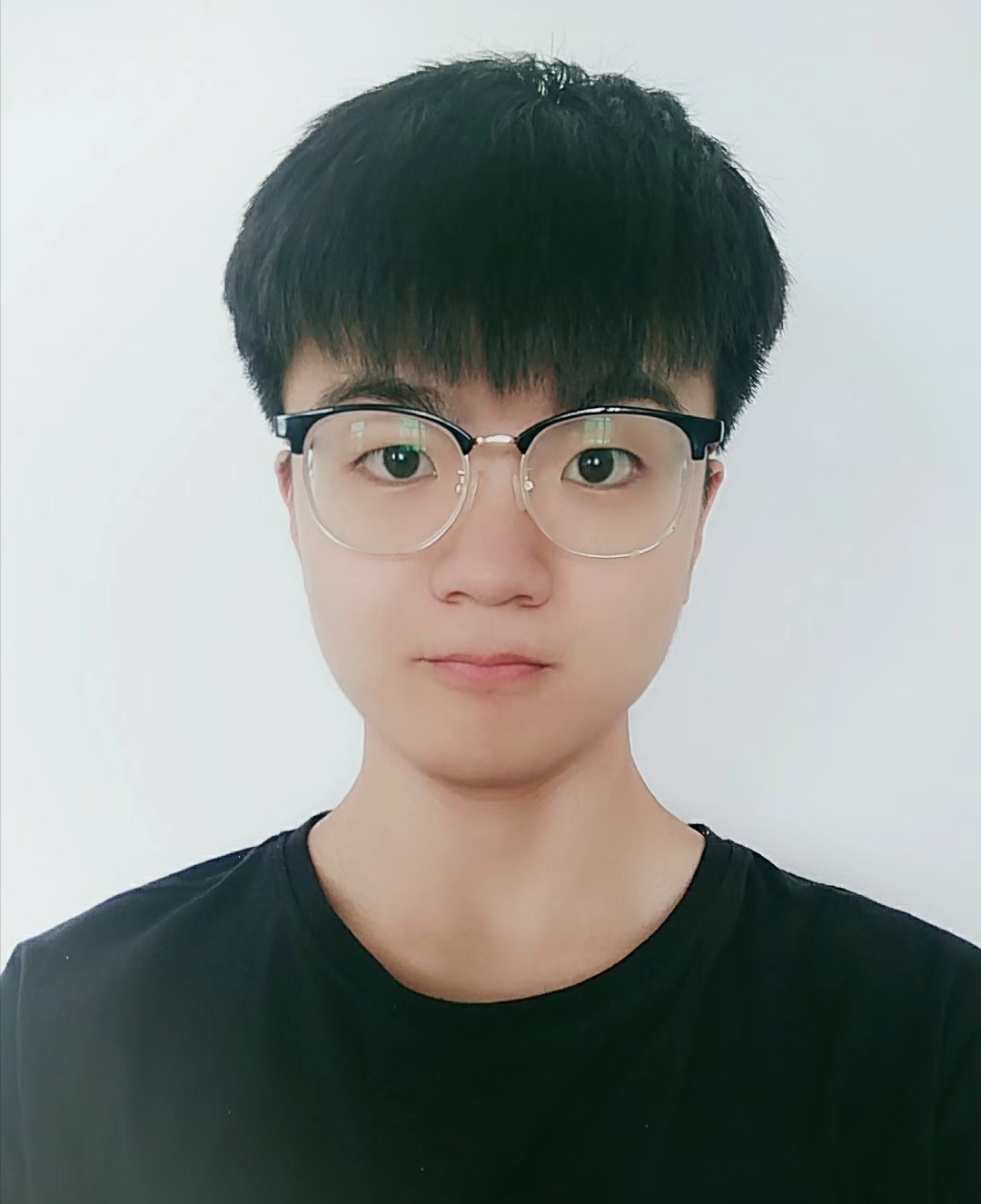}
\vspace{-10pt}
\end{figure}

\noindent{\bf Ruohao Guo} received the M.E. degree from the College of Information and Electrical Engineering, China Agricultural University, Beijing, China, in 2022. He has been working toward the Ph.D. degree with the National Key Laboratory of General Artificial Intelligence, School of Intelligence Science and Technology, Peking University, Beijing, China, since 2022. 

His research interests include computer vision and multimodal processing.

E-mail: ruohguo@stu.pku.edu.cn

\begin{figure}[h]
\centering
\includegraphics[width=0.3\textwidth]{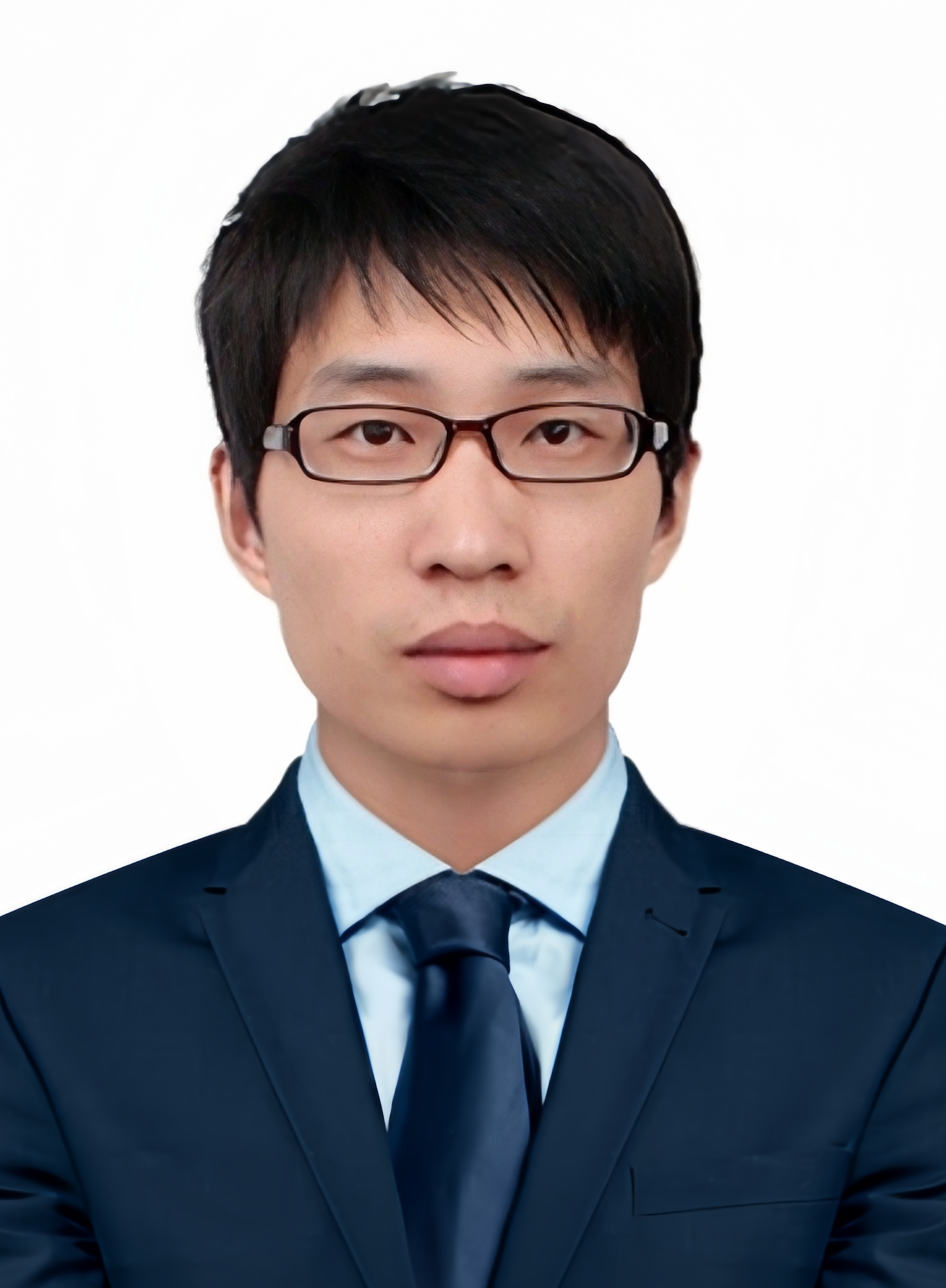}
\end{figure}

\noindent{\bf Kun Ding} received the B.S. degree in automation from Tianjin University of Science and Technology, Tianjin, China, in 2011, and the M.S. and Ph.D. degrees in pattern recognition and intelligent system from the Institute of Automation, Chinese Academy of Sciences, Beijing, China, in 2014 and 2017, respectively. He is currently an assistant professor with the State Key Laboratory of Multimodal Artificial Intelligence Systems (MAIS), Institute of Automation, Chinese Academy of Sciences. 

His research interests include pattern recognition, machine learning, and computer vision.

E-mail: kun.ding@ia.ac.cn (Corresponding author)

ORCID iD: 0000-0002-2256-8815

\begin{figure}[h]
\centering
\includegraphics[width=0.3\textwidth]{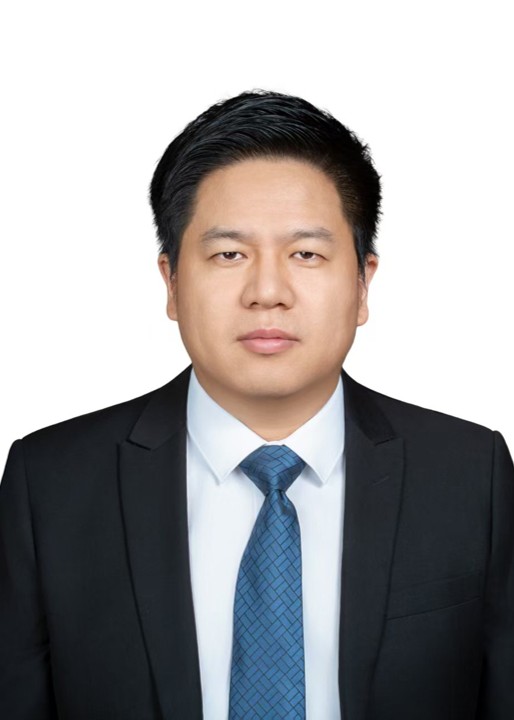}
\end{figure}

\noindent{\bf Xin Gu} received his B. S. degree in Measurement Technology and Instruments from Nanjing University of Aeronautics and Astronautics, Nanjing, China, in 2011. He is currently a Research Fellow with the Department of Research and Development Center, China Academy of Launch Vehicle technology, Beijing.

His research interests included unmanned systems and artificial intelligence.

E-mail: nync396@126.com

ORCID iD: 0009-0002-8685-8309

\begin{figure}[h]%
\centering
\includegraphics[width=0.3\textwidth]{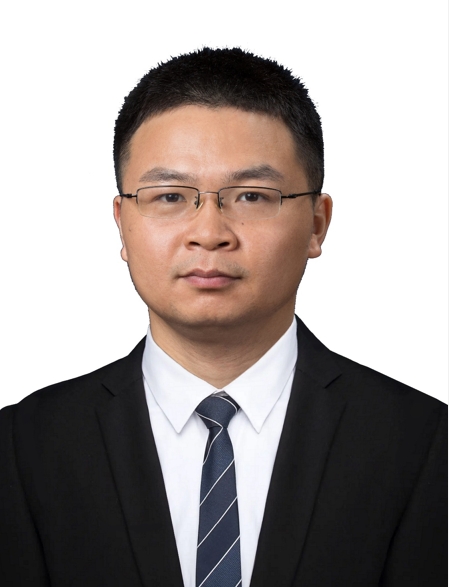}
\end{figure}

\noindent{\bf Bin Fan} received the B.S. degree from the Beijing University of Chemical Technology, Beijing, China, in 2006, and the Ph.D. degree from the National Laboratory of Pattern Recognition, Institute of Automation, Chinese Academy of Sciences, Beijing, China, in 2011. He is currently a Professor with School of Intelligence Science and Technology and the Institute of Artificial Intelligence, University of Science and Technology Beijing, China. 

He has wide research interests in computer vision, pattern recognition, image processing, and multimedia.

E-mail: bin.fan@ieee.org

ORCID iD: 0000-0002-1155-467X

\begin{figure}[h]%
\centering
\includegraphics[width=0.3\textwidth]{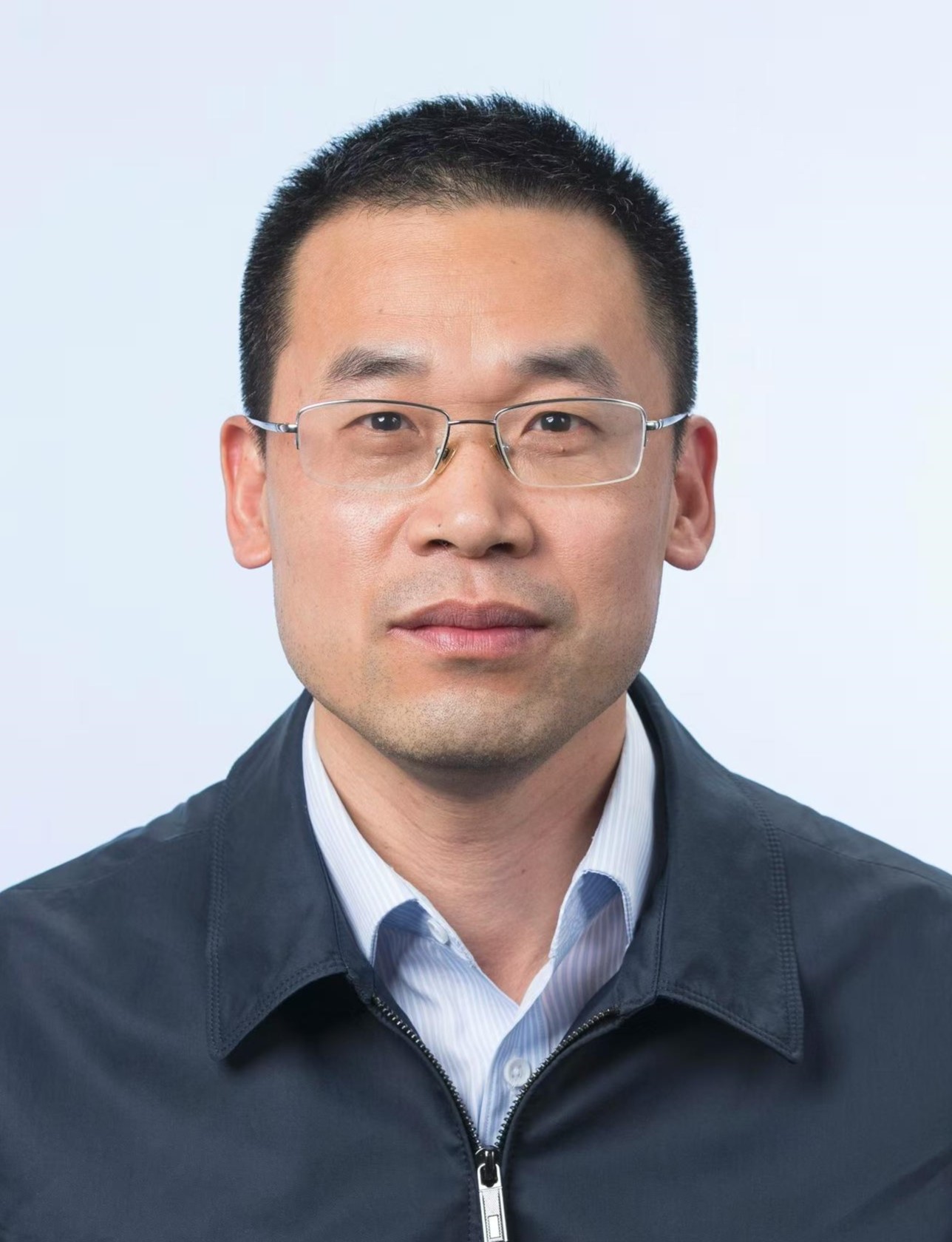}
\end{figure}

\noindent{\bf Shiming Xiang} received the B.S. degree in mathematics from Chongqing Normal University, Chongqing, China, in 1993, the M.S. degree from Chongqing University, Chongqing, China, in 1996, and the Ph.D. degree from the Institute of Computing Technology, Chinese Academy of Sciences, Beijing, China, in 2004. From 1996 to 2001, he was a Lecturer with the Huazhong University of Science and Technology, Wuhan, China. He was a Postdoctorate Candidate with the Department of Automation, Tsinghua University, Beijing, until 2006. He is currently a Professor with the State Key Laboratory of Multimodal Artificial Intelligence Systems, Institute of Automation, Chinese Academy of Sciences, Beijing. 

His research interests include pattern recognition, machine learning, and computer vision.

E-mail: smxiang@nlpr.ia.ac.cn

ORCID iD: 0000-0002-2089-9733

\end{document}